\documentclass[10pt,twocolumn,letterpaper]{article}

\usepackage{cvpr}
\usepackage{times}
\usepackage{epsfig}
\usepackage{graphicx}
\usepackage{amsmath}
\usepackage{amssymb}
\usepackage{multirow}
\usepackage{url}
\usepackage{threeparttable}
\usepackage{booktabs}
\usepackage{dcolumn}
\usepackage{graphicx}
\usepackage{float} 
\usepackage{enumitem}
\usepackage{float}
\usepackage{subfigure}
\newcolumntype{d}[1]{D{.}{.}{#1}}

\usepackage[pagebackref=false,breaklinks=true,letterpaper=true,colorlinks,bookmarks=false]{hyperref}

 \cvprfinalcopy %

\def\cvprPaperID{3147} %

\usepackage{caption}
\ifcvprfinal\fi
\begin{document}

\title{MobileFAN: Transferring Deep Hidden Representation for Face Alignment}

\author{Yang Zhao$^{1,}$$^{2}$
~~~~~Yifan Liu$^{2}$
~~~~~Chunhua Shen$^{2}$
~~~~~Yongsheng Gao$^{1}$
~~~~~Shengwu Xiong$^{3}$\\
$^{1}$Griffith University ~ ~ 
$^{2}$The University of Adelaide ~ ~ 
$^{3}$Wuhan University of Technology
}
\maketitle
\thispagestyle{empty}

\begin{abstract}
Facial landmark detection is a crucial prerequisite for many face analysis applications. Deep learning-based methods currently dominate the approach of addressing the facial landmark detection. However, such works generally introduce a large number of parameters, resulting in high memory cost. 
In this paper, we aim for a lightweight as well as effective solution to facial landmark detection. 
To this end, we propose an effective lightweight model, namely \textit{Mobile Face Alignment Network} (MobileFAN), using a simple backbone MobileNetV$2$ as the encoder and three deconvolutional layers as the decoder. The proposed MobileFAN, with only $8\%$ of the model size and lower computational cost, achieves superior or equivalent performance compared with state-of-the-art models. Moreover, by transferring the geometric structural information of a face graph from a large complex model to our proposed MobileFAN through feature-aligned distillation and feature-similarity distillation, the performance of MobileFAN is further improved in effectiveness and efficiency for face alignment.
Extensive experiment results on three challenging facial landmark estimation benchmarks including COFW, 300W and WFLW show the superiority of our proposed MobileFAN against state-of-the-art methods.
\end{abstract}



\vspace{-0.5cm}
\section{Introduction}
Facial landmark detection, a.k.a, face alignment, is a crucial step for various downstream face applications including face recognition \cite{soltanpour2017survey}, facial attributes estimation  \cite{zhuang2018multi}
, face pose estimation \cite{hong2018multi} and so forth. Face alignment aims to find the coordinates of several predefined landmarks or parts, such as eye center, eyebrow, nose tip, mouth and chin, on a face graph. Although great progress has been made on accuracy improvements in the past decades
\cite{zhang2015hierarchical,jin2016face}, approaches focusing on simple, small and lightweight networks for face alignment receive relatively much less attention.

Significant improvements via deep Convolutional Neural Networks (CNNs) have been achieved on facial landmark detection recently \cite{wu2018look,dong2018style,ValleBVB18}, even though it remains a very challenging task when dealing with faces in real-world conditions (\textit{e.g.}, faces with unconstrained large pose variations and heavy occlusions). 
In order to guarantee promising performance in face alignment benchmarks, the majority of those works are designed to adopt large backbones (\textit{e.g.}, Hourglass \cite{newell2016stacked} and ResNet-$50$ \cite{HeZRS16}), carefully designed schemes (\textit{e.g.}, a coarse-to-fine cascade regression framework \cite{SunWT13}), or adding extra face structure information (\textit{e.g.}, face boundary information \cite{wu2018look}).
Recently, neural networks with small model size,  light computation cost and high accuracy, have attracted much attention because of the need of applications on mobile devices \cite{yu2016iprivacy}.
In this paper, we aim to investigate the possibility of optimizing facial landmark detection with a simpler and smaller model. 
 We propose a plain model without bells and whistles, namely \textit{Mobile Face Alignment Network} (MobileFAN), which employs an Encoder-Decoder architecture in the form of Convolution-Deconvolution Network. In the proposed MobileFAN, MobileNetV$2$ \cite{SandlerHZZC18} is adopted as the encoder, while the decoder is constructed utilizing three deconvolutional layers. Model details are illustrated in Section \uppercase\expandafter{\romannumeral3}.

More recently, knowledge distillation (KD) has attracted much attention for its simplicity and efficiency \cite{hinton2015distilling}.
Motivated by KD and TCNN \cite{wu2018facial}, which has shown that intermediate features from deep networks are good at predicting different head poses in facial landmark detection, we further introduce knowledge transfer techniques to help the training of our proposed lightweight face alignment network.
Because our proposed MobileFAN uses several deconvolutional layers sequentially to map from the input space to the output space,
we try to transfer the useful information of intermediate feature maps from a teacher to a student. 

Inspired by \cite{li2017mimicking,romero2014fitnets}, we propose to align the deconvolutional feature maps between student models and teacher models.
Specifically, the feature map generated by the student network can be transformed to a new feature map, which needs to match the same size of the corresponding feature map generated by the teacher network. Mean squared error (MSE) is used as the loss function to measure the distance between teacher's feature map and student's new feature map. We term this scheme feature-aligned distillation, which can transfer the distribution of intermediate feature map produced by the teacher network to that of the student network.

To distill more structured knowledge information from the teacher network, inspired by \cite{liu2019structured}, we apply the feature-similarity distillation to our framework.
The similarity matrix is generated by computing cosine similarity of feature vectors. We find that the similarity matrix can be used to represent the structure information of a face image. It contains the directional knowledge between features, which can be thought of a kind of structure information. 
With the help of feature-similarity distillation, the student network is trained to make its similarity matrix similar to that of the teacher network. The illustration of our method for knowledge transfer is depicted in Fig. \ref{framework_fig} (c).

The interest of this work lies in exploring a simple, small and lightweight network that can achieve comparable or even better results than the common facial landmark detection benchmarks. To summarize, our main contributions are as follows.
\begin{itemize}
\item We propose a simple and lightweight \textit{Mobile Face Alignment Network} (MobileFAN) for face alignment. The proposed MobileFAN achieves comparable results, while having less than $8\%$ of parameters compared with the sizes of state-of-the-arts.
\item Feature-aligned distillation and feature-similarity distillation are introduced and integrated with the proposed MobileFAN, leading to further improvements in alignment accuracy.
\item Extensive experimental results on three challenging benchmark datasets demonstrate the efficiency and effectiveness of our method over state-of-the-art methods in face alignment.
\end{itemize}

\section{Related Work}
In this section, we present an overview of related work on facial landmark detection and knowledge distillation.
\subsection{Facial Landmark Detection}

\noindent\textbf{Traditional Methods.} 
Facial landmark detection has been an active topic for more than twenty years. 
Recently, cascade regression attracts a lot of attention, which focuses on learning a cascade of regressors to iteratively update the shape estimation. BurgosArtizzu \textit{et al.} \cite{Burgos-ArtizzuPD13} proposed Robust Cascade Pose Regression (RCPR) that reduces exposure to outliers by detecting occlusions explicitly and using robust shape-indexed features. Explicit Shape Regression (ESR) \cite{CaoWWS12} introduced two-level boosted regression and correlation-based feature selection.
Ren \textit{et al.} \cite{ren2014face} proposed Local Binary Features (LBF) that is computationally cheap and thus enables very fast regression on the face alignment tasks.

\noindent\textbf{CNN-based Methods.}
Apart from the above early works, deep learning-based face alignment approaches have achieved state-of-the-art performance. They can be divided into two categories, namely coordinate regression-based method and heatmap regression-based method. A coordinate regression-based method estimates the landmark coordinates vector from the input image directly. The earliest work could be dated to \cite{SunWT13}. 
Sun \textit{et al.} \cite{SunWT13} trained a three-level cascade CNN to locate the facial landmarks in a coarse-to-fine manner, and obtained promising landmark detection results.
A multi-task learning framework is proposed by Zhang \textit{et al.} \cite{ZhangLLT16} to optimize face alignment and correlated facial attributes, such as pose, expression and gender, simultaneously.
More recently, Feng \textit{et al.} \cite{FengKA0W18} proposed a new loss function, namely Wingloss, to fill the gap of a better loss function in facial coordinates regression community. It shows that Wingloss with the proposed strong data augmentation method, pose-based data balancing (PDB), could obtain better performance against widely used L2 loss.
Different from the above methods, our approach regards face alignment as a dense prediction problem.

A heatmap regression-based method generates a probability heatmap for each landmark respectively. Thanks to the development of Hourglass \cite{newell2016stacked}, heatmap regression has been successfully applied to landmark localization problems.
Yang \textit{et al.} \cite{YangLZ17} adopted a supervised face transformation based on the Hourglass to reduce the variance of the target. LAB \cite{wu2018look} utilized boundary lines to characterize the geometric structure of a face image and thus improved the detection of facial landmarks. However, both the two methods rely on the Hourglass, resulting in introducing a large number of parameters.
Valle \textit{et al.} \cite{ValleBVB18} used a simple CNN to generate heatmaps of landmark locations for a better initialization to Ensemble of Regression Trees (ERT) regressor.

By contrast, our model requires neither cascaded networks nor large backbones, leading to great reduction in model parameters and computation complexity, whilst still achieving comparable or even better accuracy.

\begin{figure*}[t]
\centering
\includegraphics[width=1.0\textwidth]{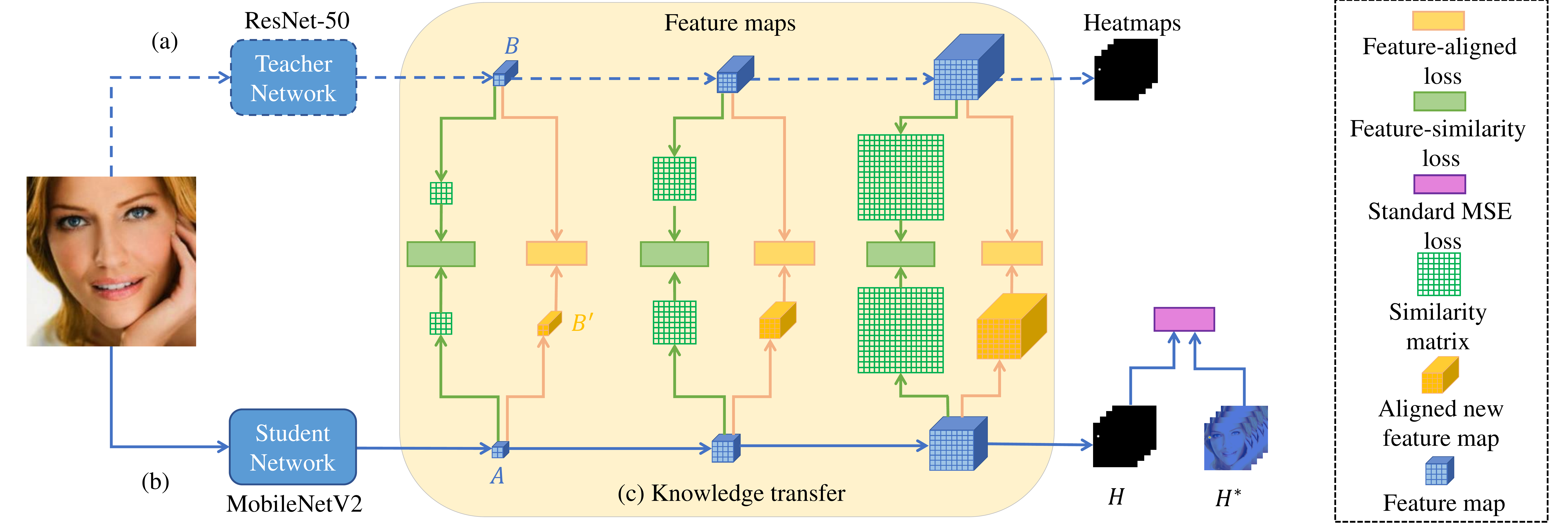}
\caption{The structure of proposed MobileFAN and an overview of our knowledge transfer framework. (a) The process indicated by the blue dotted arrow is the teacher network, which is made up of ResNet-$50$ and three deconvolutional layers. (b) The process indicated by the blue arrow is MobileFAN (the student network), which consists of MobileNetV$2$ and three deconvolutional layers. (c) Knowledge transfer module, which is designed to help the training process of MobileFAN by introducing feature-aligned distillation and feature-similarity distillation.}
\label{framework_fig}
\end{figure*}

\subsection{Knowledge Distillation}
Deep CNN models dominate the approach to solving many computer vision tasks recently \cite{hong2015multimodal, yu2019spatial}.
However, millions of parameters are commonly introduced in these Deep CNN models, leading to large model sizes and expensive computation cost. As a result, it is difficult to deploy such models to real-time applications. Therefore, it motivates researchers to focus on smaller networks that can fit large training data while maintaining the performance. Recently, knowledge distillation (KD) \cite{hinton2015distilling} has attracted much attention due to its capability of transferring rich information from a large and complex teacher network to a small and compact student network. It is widely used in model compression. Originally, KD is used in the task of image classification \cite{hinton2015distilling}, where a compact model can learn from the output of a large model, namely soft target. So the student is supervised by both softened labels and hard labels simultaneously.

Following \cite{hinton2015distilling}, some subsequent works have tried to transfer intermediate representations of the teacher network to that of the student network, and achieved great progress in image classification \cite{romero2014fitnets,zagoruyko2016paying}, object detection \cite{li2017mimicking},
and semantic segmentation \cite{liu2019structured}. Romero \textit{et al.} proposed a FitNet \cite{romero2014fitnets} which directly aligns full feature maps of the teacher model and student model. Attention transfer (AT) \cite{zagoruyko2016paying} was proposed to regularize the learning of the student network by imitating the attention maps of a powerful teacher network. 
Liu \textit{et al.} \cite{liu2019structured} proposed to distill the pair-wise information from the teacher model to a student model via the last convolution features. Unlike previous approaches, we perform the distillation through multiple features.

Given the effectiveness of knowledge distillation in the above applications, we are motivated to simultaneously use both feature-aligned distillation and feature-similarity distillation on intermediate features in this work. 

\vspace{-0.2cm}

\section{Method}
In this section, we start with an introduction of network architectures. Then we take a look at standard MSE loss and two knowledge distillation schemes: feature-aligned distillation and feature-similarity distillation.

\subsection{Network Architectures}

\noindent\textbf{Teacher Network.}
Following \cite{xiao2018simple}, we introduce a teacher network which employs an Encoder-Decoder architecture in the form of Convolution-Deconvolution Network.
The encoder is a ResNet$50$ where the last average pooling layer and the classification layer are both removed.
The decoder consists of three deconvolutional layers, with the dimension of $256$ and $4 \times 4$ kernel for each layer. 
The stride of each deconvolutional layer in the decoder is set to be $2$. Then, a $1\times1$ convolutional layer is added after the decoder to generate likelihood heatmaps.
The architecture of the teacher network is shown in Fig. \ref{framework_fig} (a).

\noindent\textbf{Student Network.}
To make our framework easy to be reconstructed, we present a student network using a similar Encoder-Decoder structure as adopted in the teacher network.
The encoder is a MobileNetV$2$ \cite{SandlerHZZC18} with the last three layers (\textit{i.e.}, one global average pooling layer and two convolutional layers) being removed. The decoder is composed of three deconvolutional layers, and is added over the last bottleneck of the MobileNetV$2$.
Each deconvolutional layer has $128$ filters with $2\times2$ kernel.
The stride of each deconvolutional layer in the student network is $2$.
Same as in \cite{newell2016stacked, xiao2018simple}, a $1\times1$ convolutional layer is added after the decoder to generate likelihood heatmaps $\textbf{H}=\{\textbf{H}_l \}_{l=1}^L$ for $L$ facial landmarks. 
An illustration of the detailed student network structure is shown in Fig. \ref{framework_fig} (b).

\subsection{Loss Function} 

\noindent\textbf{Mean Squared Error.}
Same as \cite{xiao2018simple}, we employ Mean Squared Error (MSE) loss to compare the predicted heatmaps $\textbf{H}$ and the ground-truth heatmaps $\textbf{H}^*$ generated from the annotated $2$D facial landmarks.

Specifically, $\textbf{H}^* = \{\textbf{H}_l^*\}$ is a set of $L$ response maps, one per facial landmark, where $\textbf{H}_l^*\in \mathbb{R}^{64\times64}$, $l\in\{1 \dots L\}$. 
Here heatmap $\textbf{H}_l^*$ for the $l^{th}$ landmark is made up of a $2$D gaussian centered on the landmark location.
Let $x\in \mathbb{R}^2$ be the ground-truth position of the $l^{th}$ facial landmark, the value at location $p\in \mathbb{R}^2$ in $\textbf{H}_l^*$ is defined as:

\begin{equation}
\textbf{H}_l^*(p)=\exp{(-\frac{\|p-x\|_2^2}{2\sigma^2})}.
\end{equation}

Therefore, the loss between the predicted heatmaps $\textbf{H}$ and ground-truth heatmaps $\textbf{H}^*$ is defined as:

\begin{equation}
\mathcal{L}(\textbf{H},\textbf{H}^*) =\|\textbf{H}-\textbf{H}^*\|^2.
\label{mseloss}
\end{equation}

After training, the $l^{th}$ landmark location can be generated from the corresponding predicted heatmap $\textbf{H}_l$ by transforming the highest heatvalued location from $1/4$ to the original image space.

\noindent\textbf{Knowledge Transfer.}
In the student-teacher framework, apart from the standard MSE loss, $\mathcal{L}(\textbf{H},\textbf{H}^*)$, we further introduce knowledge transfer loss to help the training of our student network (MobileFAN). 
In other words, we want the student network to learn not only the information provided by the ground-truth labels, but also the finer structure knowledge encoded by the teacher network.
Let $\mathcal{T}$, $\mathcal{S}$ and $\textbf{W}_{\mathcal{T}}$,$\textbf{W}_{\mathcal{S}}$ denote the teacher network, the student network and their corresponding weights. Details of knowledge distillation are described below.

\textbf{Feature-Aligned Distillation.}
In order to transfer richer facial details (\textit{e.g.}, exaggerated expressions and head poses) learned by a teacher network to the student network, we perform feature-aligned distillation such that the distribution of a feature of the student network is similar to that of the teacher network. Feature-aligned distillation is designed to align the feature map between a student network and a teacher network. 
Given a feature map $\textbf{A} \in \mathbb{R}^{C \times H \times W}$ from the student network $\mathcal{S}$ and a feature map $\textbf{B} \in \mathbb{R}^{C^{'} \times H \times W}$ from the teacher network $\mathcal{T}$, where $C$ and $C^{'}$ represent channels and $H\times W$ is the spatial dimensions. 
To perform the feature-aligned distillation, $\textbf{A}$ and $\textbf{B}$ should have the same size (\textit{i.e.}, $C=C^{'}$ in our case). Therefore, we firstly adopt a $1 \times 1$ convolutional layer to take an input of feature map $\textbf{A}$ and output a new feature map $\textbf{B}^{'} \in \mathbb{R}^{C^{'} \times H \times W}$ that has the same size of the feature map $\textbf{B}$ in the teacher network. Therefore, the feature-aligned transfer loss between the teacher network and student network is defined as:

\begin{equation}
\mathcal{L}_{FA} = \|\textbf{B}-\textbf{B}^{'}\|^2.
\end{equation}

\textbf{Feature-Similarity Distillation.}
Facial images are geometrically constrained. As illustrated before, we adopt feature-similarity distillation to transfer more structural information from the teacher network to the student network (MobileFAN) by comparing their similarity matrix. The similarity matrix represents the basic facial structures and textures, which can provide richer directional information to facial landmark detection.
We perform cosine similarity computation on the whole feature map, making the relative spatial positions between facial landmarks more precisely.
Given a feature map that has dimensions of $C \times H \times W $, where $C$ is the total
number of channels and $H \times W$ is the feature map size,
$\textbf{f}_i \in \mathbb{R}^C$ denotes a feature vector extracted from the $i^{th} (i=1,...,H \times W$) spatial location of this feature map.
Therefore, the cosine similarity $a_{ij}$
between the $i^{\rm{th}}$ feature vector $\textbf{f}_i$ and $j^{\rm{th}}$ feature vector $\textbf{f}_j$ is calculated as:

\begin{equation}
a_{ij} =  \frac{\textbf{f}_i \cdot \textbf{f}_j}{\|\textbf{f}_i \| \|\textbf{f}_j \|}.
\end{equation}

Let $a_{ij}^s$ denote the similarity between the $i^{\rm{th}}$ feature vector and $j^{\rm{th}}$ feature vector computed from the feature map $\textbf{A}$, and $a_{ij}^t$ denote the similarity between the $i^{\rm{th}}$ feature vector and $j^{\rm{th}}$ feature vector computed from the feature map $\textbf{B}$. The feature-similarity transfer loss is then formulated as:

\begin{equation}
\mathcal{L}_{FS} = \frac{\sum\limits_{i \in K} \sum\limits_{j \in K} (a_{ij}^s - a_{ij}^t)^2} 
{(H \times W)^2},
\end{equation}
where $K = \{1,2, \dots, H \times W\}$ denotes all the locations.

\subsection{Distillation over Scales}
In order to transfer low-, mid- and high-level useful geometric structure information form the teacher network to the student network (MobileFAN), we extend the combination of the feature-aligned distillation and feature-similarity distillation to three deconvolutional layers.
As shown in Fig. \ref{framework_fig} (c), three deconvolutional layers of the student network are guided by that of the teacher network, where significantly richer facial details are provided. It is different from most previous methods \cite{liu2019structured} that only add supervision on the last convolutional layer.
Then we perform both feature-aligned distillation and feature-similarity distillation on the three deconvolutional feature maps during training.
So the student network is trained to optimize the following loss function:

\begin{equation}
\mathcal{L}_{KD} = \mathcal{L}(\textbf{H},\textbf{H}^*) + \lambda\sum_{r=1}^3(\mathcal{L}_{FA}^r + \mathcal{L}_{FS}^r),
\label{KD_loss}
\end{equation}
where $\lambda$ is a tunable parameter to balance the MSE loss and the distillation loss.
$\mathcal{L}_{FA}^r$ and $\mathcal{L}_{FS}^r$ are the feature-aligned loss and feature-similarity loss of the $r^{\rm{th}}$ deconvolutional layer.
Extensive experiments show that with the help of distilled knowledge from different feature map scales, the performance of facial landmark detection can be significantly increased.

\subsection{Learning Procedure}

\noindent\textbf{Training the proposed MobileFAN.}
To evaluate the performance of the proposed \textit{vanilla} MobileFAN, we optimize MobileFAN only with the standard MSE loss (as illustrated in Equation (\ref{mseloss})) without any extra losses. The experimental results indicate that our proposed MobileFAN, a simple and small network, still can handle the problem of facial landmark detection with satisfying performance.

\noindent\textbf{Training MobileFAN with distilled knowledge.}
To transfer the distilled knowledge from a large complicated network to the proposed MobileFAN, we regard MobileFAN as a student network in a student-teacher framework. Fig. \ref{framework_fig} summarizes the training of the knowledge transfer framework. Specifically, a teacher network (Fig. \ref{framework_fig} (a)) is pre-trained and the parameters are kept frozen during training.
The training stage of the proposed MobileFAN is supervised by standard MSE loss, feature-aligned loss and feature-similarity loss. In other words, guided by the pre-trained parameters $\textbf{W}_{\mathcal{T}}$ of the teacher network, we train the parameters of the MobileFAN $\textbf{W}_{\mathcal{S}}$ to minimize Equation (\ref{KD_loss}). 

\section{Experiments}
\subsection{Datasets}
We perform experiments on three challenging public datasets:
the Caltech Occluded
Faces in the Wild (COFW) dataset \cite{Burgos-ArtizzuPD13}, the $300$ Faces in the Wild ($300$W) dataset \cite{SagonasTZP13} and the Wider Facial Landmarks in the Wild (WFLW)
dataset \cite{wu2018look}.

\begin{figure}[ht]
\centering
\subfigure[COFW]{
\label{dataset_fig_a}
\includegraphics[width=0.12\textwidth]{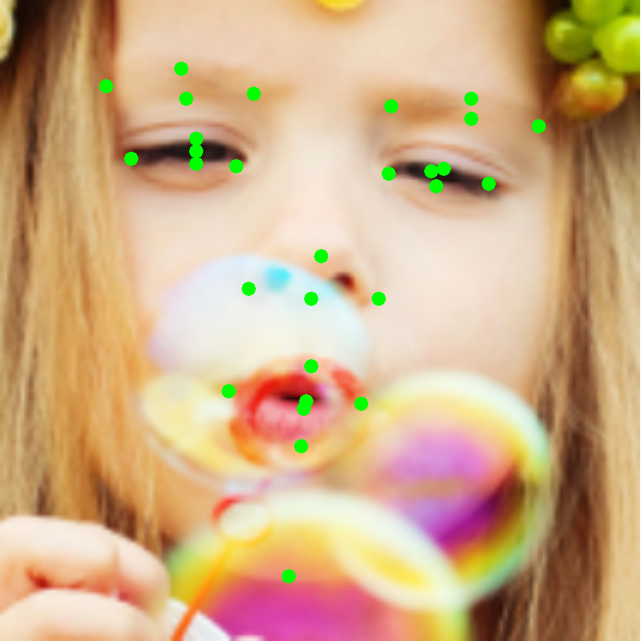}} 
\subfigure[$300$W]{
\label{dataset_fig_b}
\includegraphics[width=0.12\textwidth]{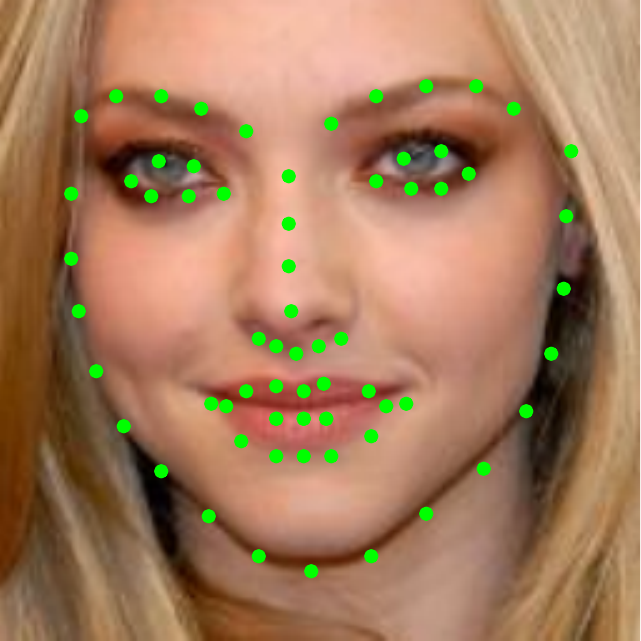}}
\subfigure[WFLW]{
\label{dataset_fig_c}
\includegraphics[width=0.12\textwidth]{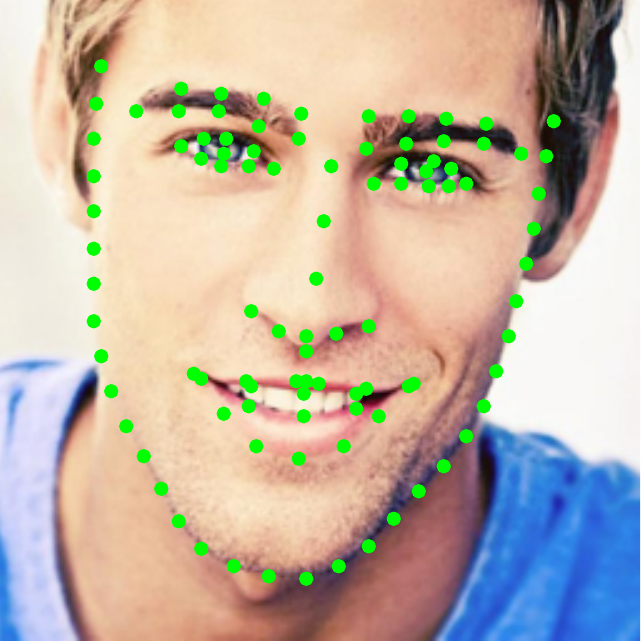}}
\vspace{-0.3cm}
\caption{An illustration of the landmark annotations of (a) COFW dataset, (b) $300$W dataset and (c) WFLW dataset.}
\label{dataset_fig}
\end{figure}

\noindent\textbf{COFW.}
The face images in COFW comprise heavy occlusions and large shape variations, which are common issues in realistic conditions \cite{Burgos-ArtizzuPD13}. Its training set has $1345$ faces, and the testing set has $507$ faces. Each image in the COFW dataset has $29$ manually annotated landmarks, as shown in Fig. \ref{dataset_fig_a}.

\noindent\textbf{300W.}
The $300$W \cite{SagonasTZP13} dataset is a widely used facial landmark detection benchmark, which consists of HELEN, LFPW, AFW and IBUG datasets. Images in HELEN, LFPW and AFW datasets are collected in the wild environment, where large pose variations, expression variations, and partial occlusions may exist.
There are $68$ annotated facial landmarks in each face from $300$W dataset, as shown in Fig. \ref{dataset_fig_b}. We follow the same protocol as used in \cite{RenCWS16} to adopt $3148$ images for training ($2000$ images from the training subset of HELEN dataset, $811$ images from the training subset of LFPW dataset and $337$ images from the full set of AFW dataset). 
For testing, the Full test set has $689$ images including Common subset ($554$ images) and Challenging subset ($135$ images). Here Common subset is composed of HELEN test subset ($330$ images) and LFPW test subset ($224$ images), while Challenging subset is the IBUG dataset.

\noindent\textbf{WFLW.}
WFLW \cite{wu2018look} is a recently proposed facial landmark dataset based on WIDER FACE. It comprises $7500$ face images (for training) and $2500$ face images (for testing) with $98$ manual annotated landmarks (shown in Fig. \ref{dataset_fig_c}), respectively. Faces in WFLW are collected under unconstrained conditions, such as large variations in poses, exaggerated expressions and heavy occlusions.
To validate the robustness against each different condition, WFLW is further divided into several subsets including large pose ($326$ images), expression ($314$ images),
illumination ($698$ images), make-up ($206$ images), occlusion ($736$ images) and blur ($773$
images). We report the results of all the competing methods on the whole test set and each testing subset in the WFLW dataset.
\vspace{-0.1cm}

\begin{table}[t]
\setlength{\tabcolsep}{2.5pt}
\scriptsize
\centering
\caption{A summary of the evaluation protocols used in our experiments.}
\begin{tabular}{  l|c|c|c|c}
\hline
 Protocol & Training Set Size & Test Set Size & \#Landmarks & Normalisation Term \\
\hline
COFW & $1,345$ & $507$  & $29$ & inter-ocular distance\\
$300$W Full set & $3,148$ & $689$ & $68$ & inter-ocular distance\\ 
WFLW & $7,500$  & $2,500$  & $98$ & inter-ocular distance\\
\hline
\end{tabular}
\label{evaluation_protocols}
\vspace{-0.5cm}
\end{table}

\vspace{-0.5cm}
\subsection{Evaluation Metrics and Implementation Details}
\noindent\textbf{Evaluation Metrics.}
We adopt the normalized mean error and the area-under-the curve (AUC) as metrics for evaluation. For all the datasets (COFW, $300$W and WFLW), we use the distance between the outer eye corners (inter-ocular distance) as the
normalization term \cite{Burgos-ArtizzuPD13,wu2018look}.

The mean error is defined as the average Euclidean distance between the predicted facial landmark
locations $p_{i,j}$ and their corresponding ground-truth facial landmark locations $g_{i,j}$:

\begin{equation}
error = \frac{1}{N}\sum_{i=1}^L\frac{\frac{1}{L}\sum_{j=1}^L|p_{i,j}-g_{i,j}|_2^2}{d}.
\end{equation}
where $N$ is the number of images in the test set, and $L$ is the number of landmarks (as illustrated in TABLE \ref{evaluation_protocols}). $d$ is the normalization factor, which is the inter-ocular distance.

We also measure the Cumulative Errors Distribution (CED) curve and the failure rate (which is defined as the proportion of failed detected faces) on these benchmarks. Specifically, any normalized error above $0.1$ is considered as a failure \cite{wu2018look}. The summary of detailed evaluation protocols used in our experiments is listed in TABLE \ref{evaluation_protocols}.

\noindent\textbf{Implementation Details.}
All the face images including both training and testing images are cropped and scaled to $256 \times 256$ according to center location and provided bounding box \cite{wu2018look,chen2019adversarial}. Standard data augmentation is
performed to make networks robust to data variations. Specifically, we follow \cite{FengKA0W18,wu2018look} to augment
samples by ($\pm30$ degree) in-plane rotation, ($0.75$-$1.25$) scaling and
randomly horizontal flip with the probability of $50\%$. In the training, Adam optimizer is used with a mini-batch size of $8$ for $80$ epochs. The base learning rate is $10^{-3}$, and it drops to $10^{-4}$ at $30$th epochs and $10^{-5}$ at $50$th epochs respectively. $\lambda$ is set to be $10^{-2}$ for COFW dataset and $10^{-4}$ for $300$W and WFLW dataset. For implementation, all our face estimation models are trained with Pytorch $0.4.0$ toolbox on one GPU.

\begin{table}[t]
\setlength{\tabcolsep}{2.5pt}
\scriptsize
\centering
\caption{Mean error (\%) and failure rate (\%) on COFW test set.}
\vspace{-0.2cm}
\begin{tabular}{  l|c |c }
\hline
Method & Mean Error & Failure Rate  \\
\hline
Human  \cite{Burgos-ArtizzuPD13} &$5.6$ & -\\
RCPR \cite{Burgos-ArtizzuPD13} & $8.50$ & $20.00$ \\
HPM \cite{GhiasiF14}& $7.50$ & $13.00$ \\
CCR \cite{FengHKCW15Cascaded}   & $7.03$ & $10.9$\\
DRDA \cite{ZhangKSC16} & $6.46$ & $6.00$ \\
RAR \cite{XiaoFXLYK16} & $6.03$ & $4.14$ \\
SFPD \cite{WuGJ17} & $6.40$ & -\\
DAC-CSR \cite{FengKC0W17}  & $6.03$ & $4.73$ \\
CNN$6$ (Wing + PDB) \cite{FengKA0W18}& $5.44$ & $3.75$\\
ResNet$50$ (Wing + PDB) \cite{FengKA0W18}& $5.07$ & $3.16$\\
LAB \cite{wu2018look} & $3.92$ & $0.39$ \\
\hline
Teacher  & $3.65$ & $0.59$ \\
MobileFAN ($0.5$)& $4.01$ & $0.99$ \\
MobileFAN  & $\mathbf{3.82}$ & $0.59$\\
\hline
MobileFAN ($0.5$) + KD & $\mathbf{3.68}$ & $0.59$ \\
MobileFAN + KD   & $\mathbf{3.66}$ & $0.59$ \\
\hline
\end{tabular}
\label{cofw_testset}
\vspace{-0.5cm}
\end{table}

\subsection{Comparison with state-of-the-art methods}
We compare the proposed method against the state-of-the-art methods on each dataset. To further explore whether the effectiveness of a smaller decoder on face alignment task, we adopt a channel-halved version of MobileFAN, where we use $64$ dimension to replace $128$ dimension of each deconvolutional layer. This architecture is denoted as MobileFAN ($0.5$). We apply our distillation method to both of the two lightweight networks: MobileFAN and MobileFAN ($0.5$). For simplicity, we name our full models trained using the combination of feature-aligned distillation and feature-similarity distillation of all the deconvolutional layers to be ``MobileFAN + KD" and ``MobileFAN ($0.5$) + KD". Similarly, the baseline models, MobileFAN without distillation and MobileFAN ($0.5$) without distillation, are named as ``MobileFAN" and ``MobileFAN ($0.5$)". We use ``Teacher" to represent our proposed teacher network.

\begin{figure}[ht]
\centering
\includegraphics[width=8cm]{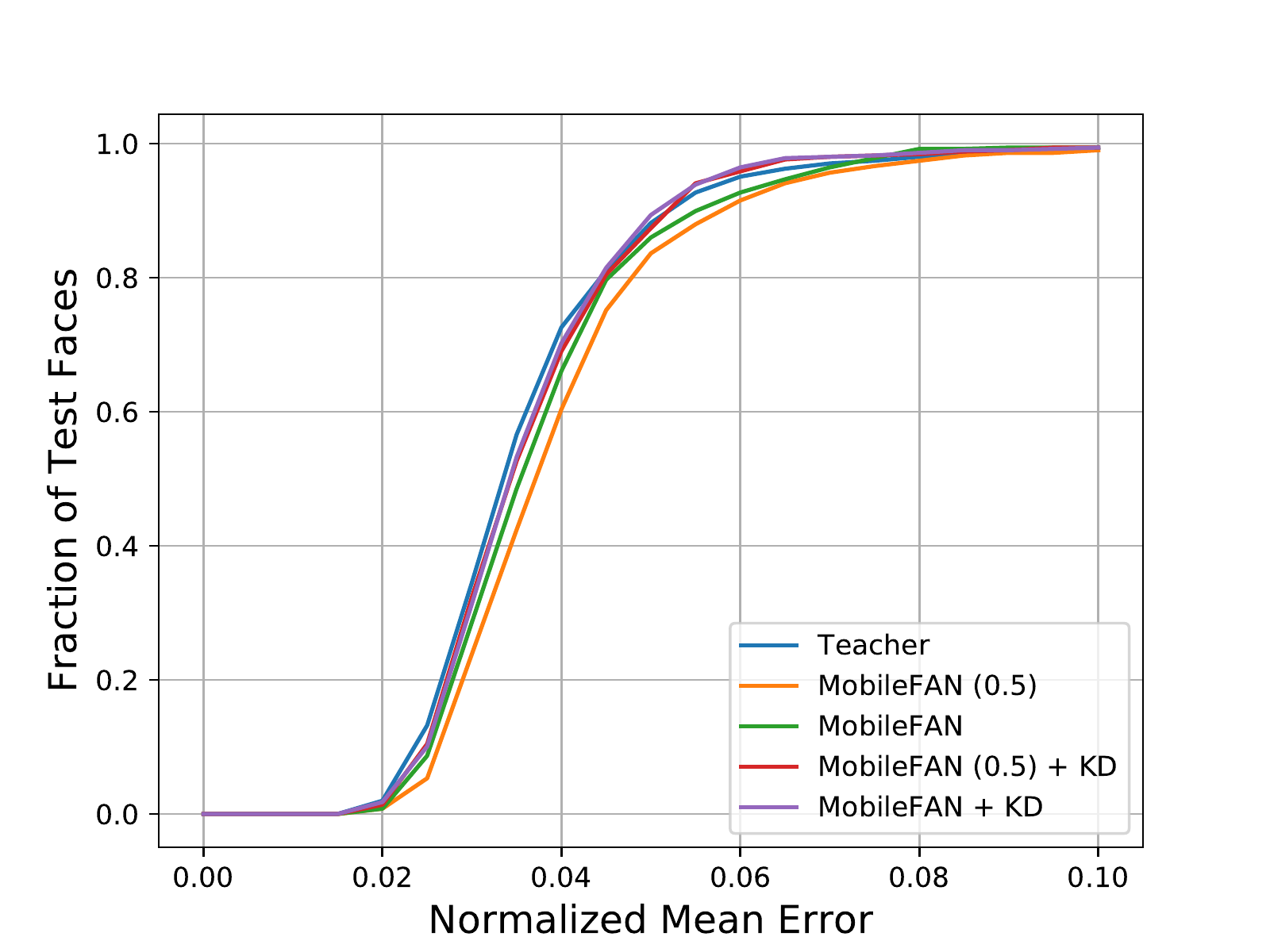}
\vspace{-0.3cm}
\caption{CED curves of the baselines, teacher network and the proposed MobileFAN with KD on COFW test set.}
\label{cofw_ced}
\end{figure}

\begin{figure}[ht] 
\centering
\includegraphics[width=0.1\textwidth]{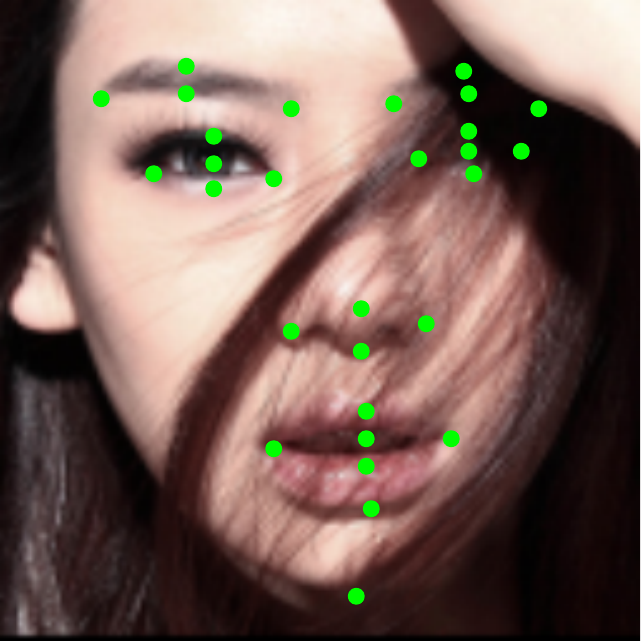}
\includegraphics[width=0.1\textwidth]{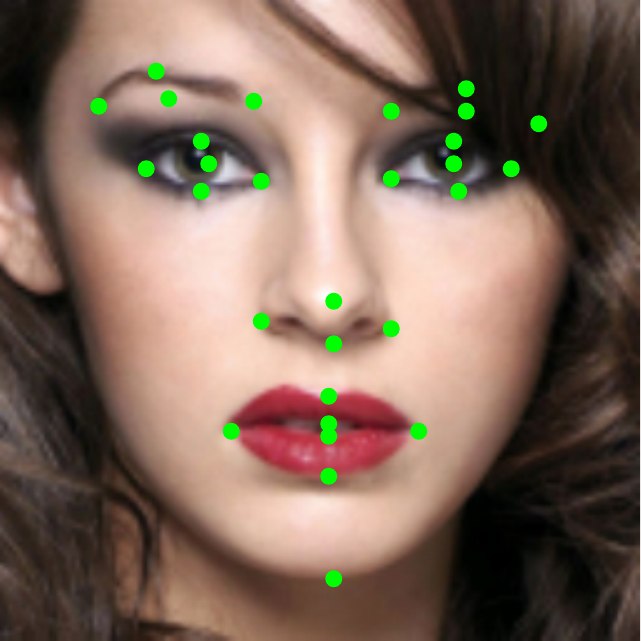}
\includegraphics[width=0.1\textwidth]{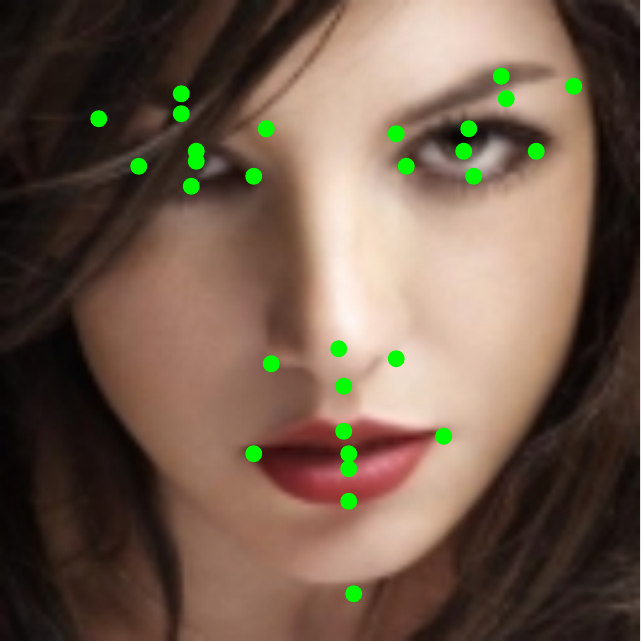}
\includegraphics[width=0.1\textwidth]{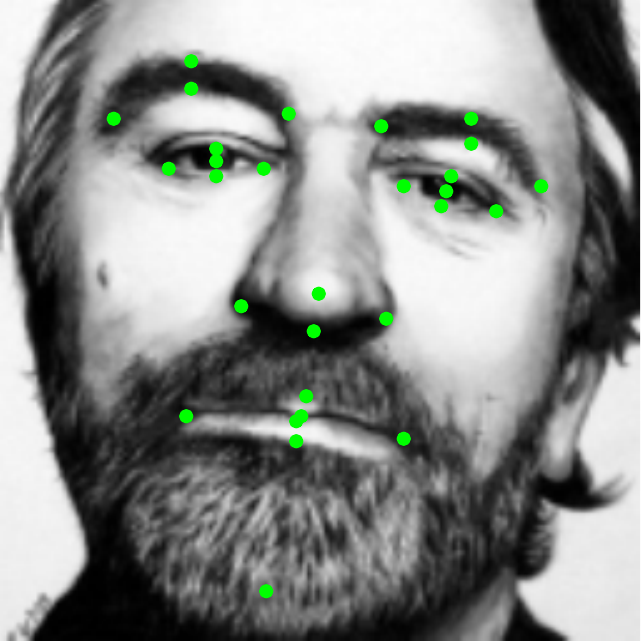}

\includegraphics[width=0.1\textwidth]{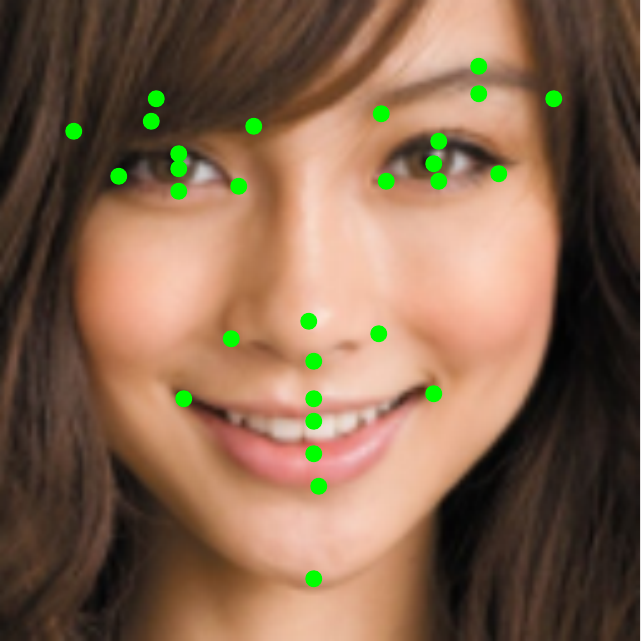}
\includegraphics[width=0.1\textwidth]{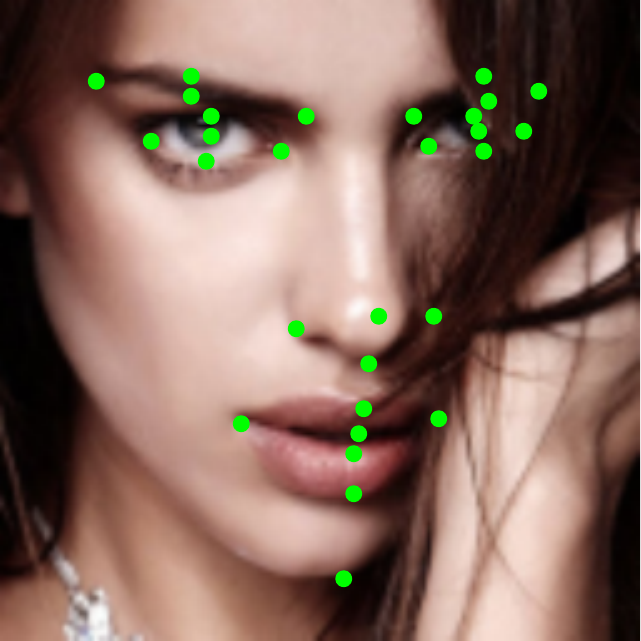}
\includegraphics[width=0.1\textwidth]{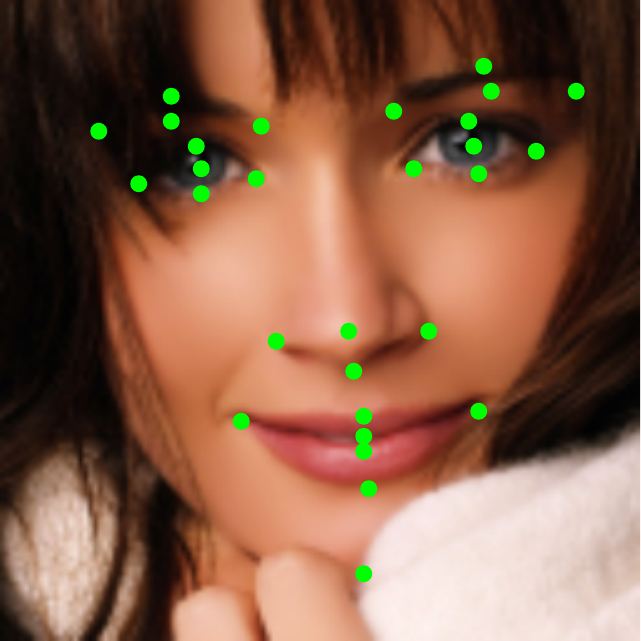}
\includegraphics[width=0.1\textwidth]{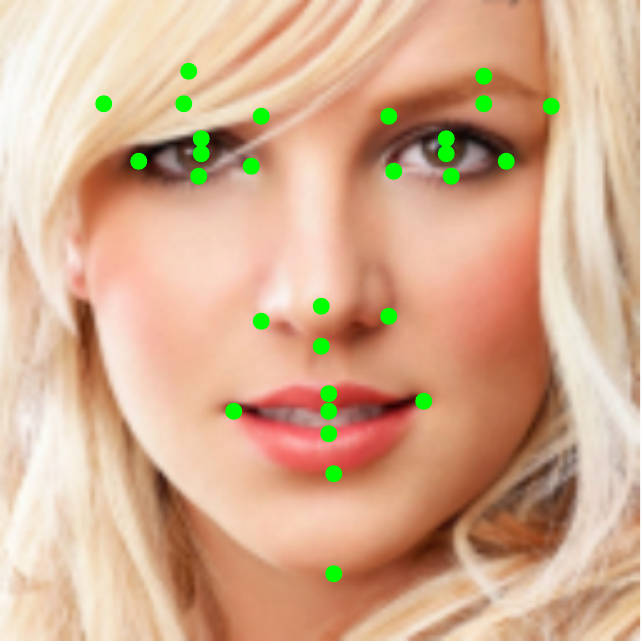}

\includegraphics[width=0.1\textwidth]{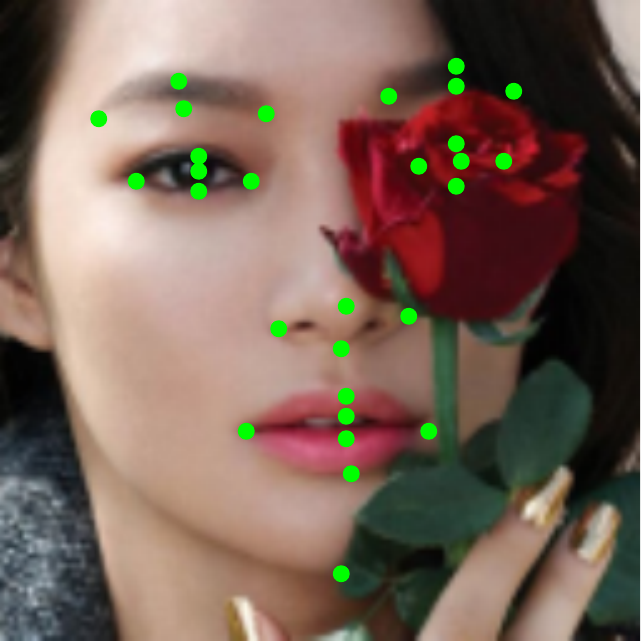}
\includegraphics[width=0.1\textwidth]{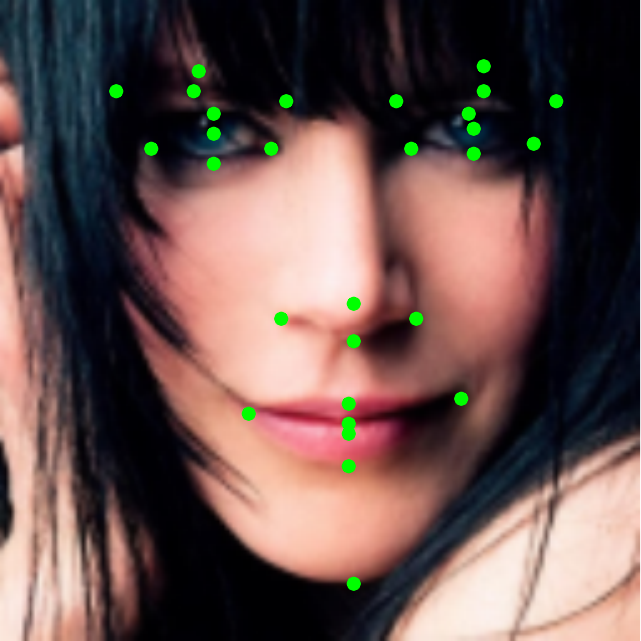}
\includegraphics[width=0.1\textwidth]{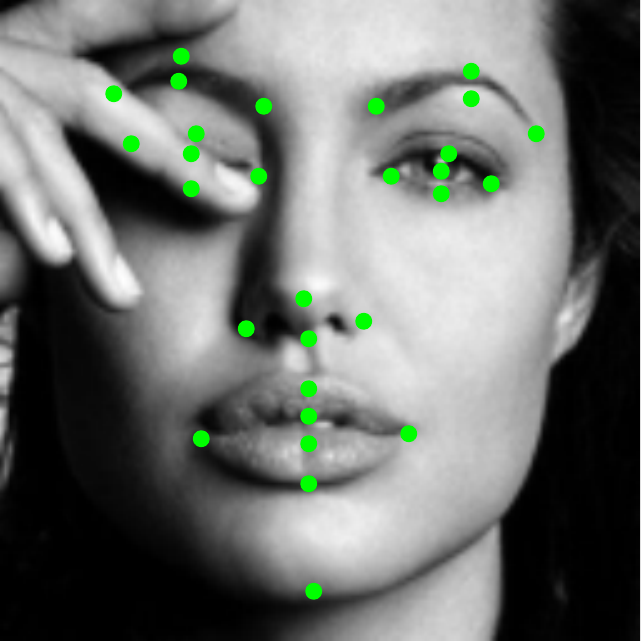}
\includegraphics[width=0.1\textwidth]{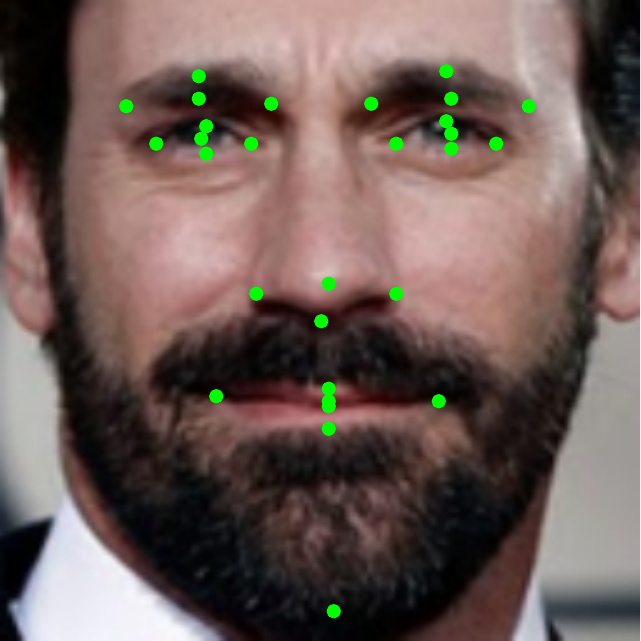}

\vspace{-0.2cm}
\caption{Example alignment results of MobileFAN on COFW \cite{Burgos-ArtizzuPD13} test set.}
\label{cofw_results}
\vspace{-0.6cm}
\end{figure}

\noindent\textbf{Evaluation on COFW.}
In TABLE \ref{cofw_testset}, we provide the results of the state-of-the-art methods in COFW test set.
We can see that the proposed simple and small ``MobileFAN" achieves $3.82\%$ mean error with $0.59\%$ failure rate without any extra information.
Although the mean error of ``MobileFAN ($0.5$)" is a little higher than that of LAB \cite{wu2018look}, it is still a comparable result ($4.01\%$ mean error with $0.99\%$ failure rate).
With the knowledge transferred from ``Teacher", our method outperforms existing methods with a margin of over $0.24\%$ in mean error reduction. More specifically, our proposed models, ``MobileFAN + KD" and ``MobileFAN ($0.5$) + KD", achieve the best performances on COFW dataset, with about $6.63\%$ and $6.12\%$ improvements in mean error reduction over LAB \cite{wu2018look} with extra boundary information. ``MobileFAN + KD" achieves the comparable result ($3.66\%$ mean error with $0.59\%$ failure rate) to the teacher network ($3.65\%$ mean error with $0.59\%$ failure rate). This is not surprising since our proposed distillation method provides rich structural information of a face image, which may contribute to the performance of facial landmark detection. The CED curves in Fig. \ref{cofw_ced} show that the distilled small networks gain better performance than baselines as well as achieve comparable performance to the ``Teacher" network. Fig. \ref{cofw_results} shows some example alignment results, demonstrating the effectiveness on various occlusions of ``MobileFAN".

\begin{figure}[ht] 
\centering
\vspace{-0.3cm}
\includegraphics[width=0.1\textwidth]{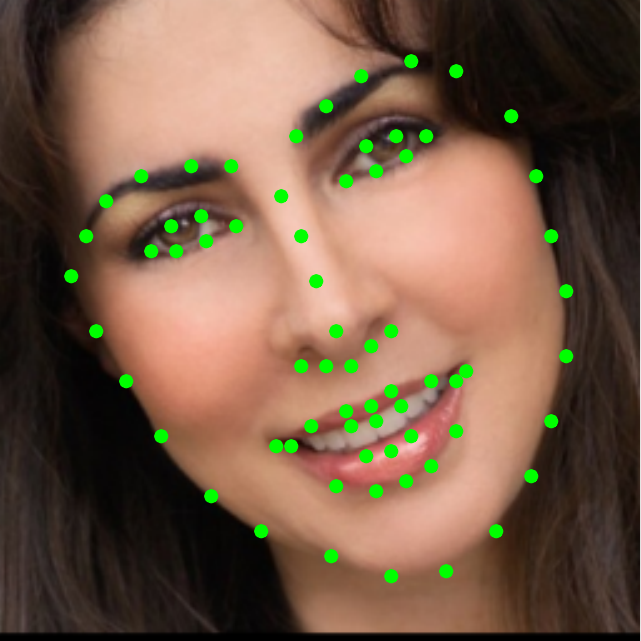}
\includegraphics[width=0.1\textwidth]{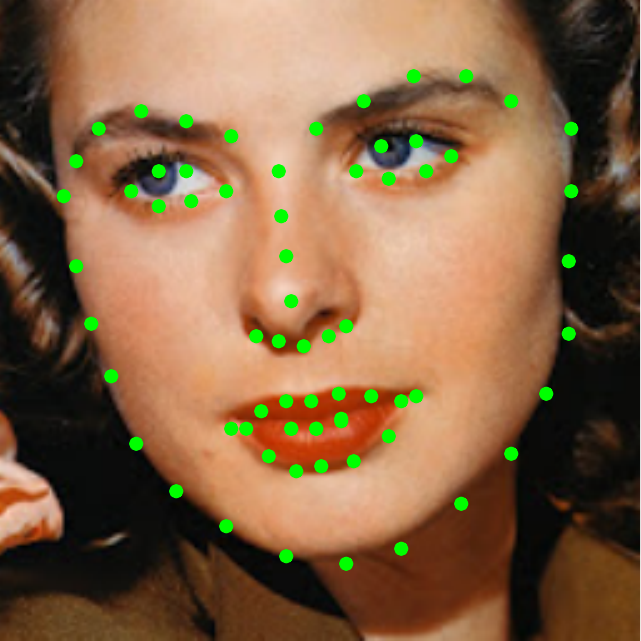}
\includegraphics[width=0.1\textwidth]{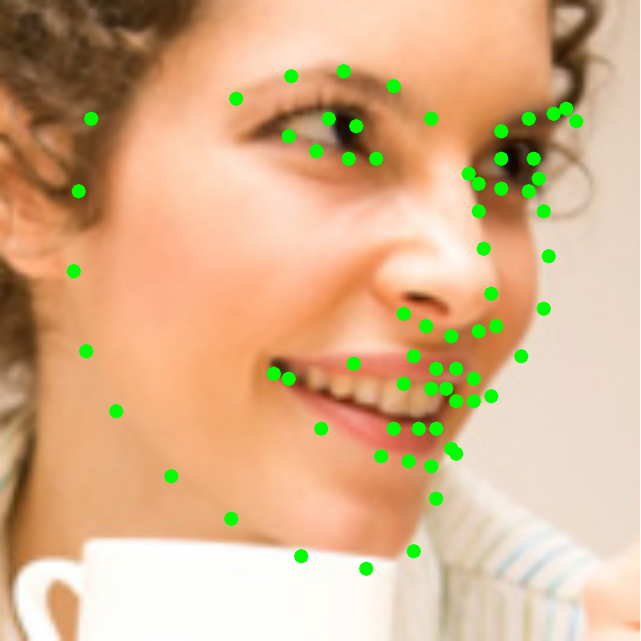}
\includegraphics[width=0.1\textwidth]{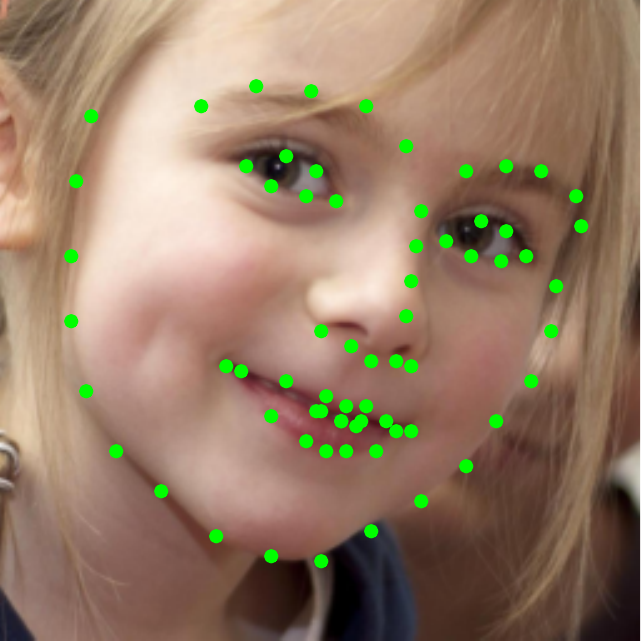}

\includegraphics[width=0.1\textwidth]{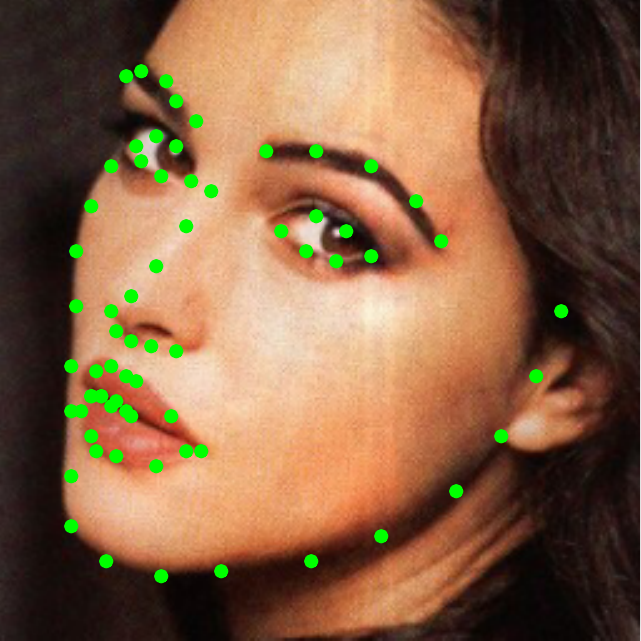}
\includegraphics[width=0.1\textwidth]{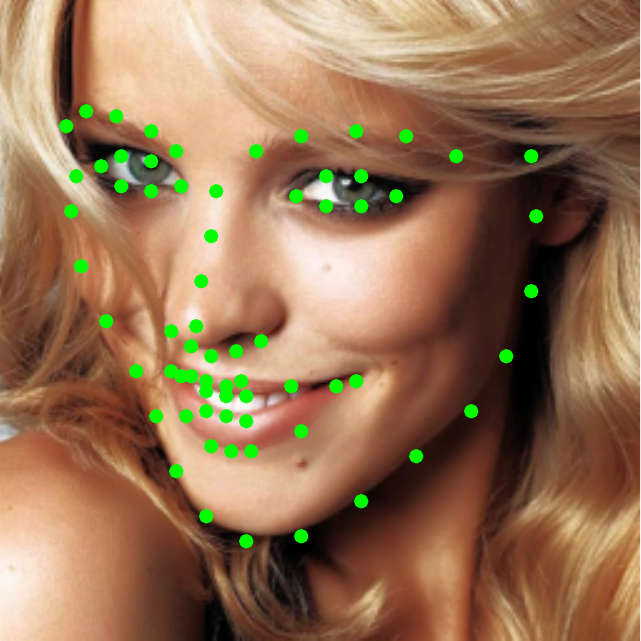}
\includegraphics[width=0.1\textwidth]{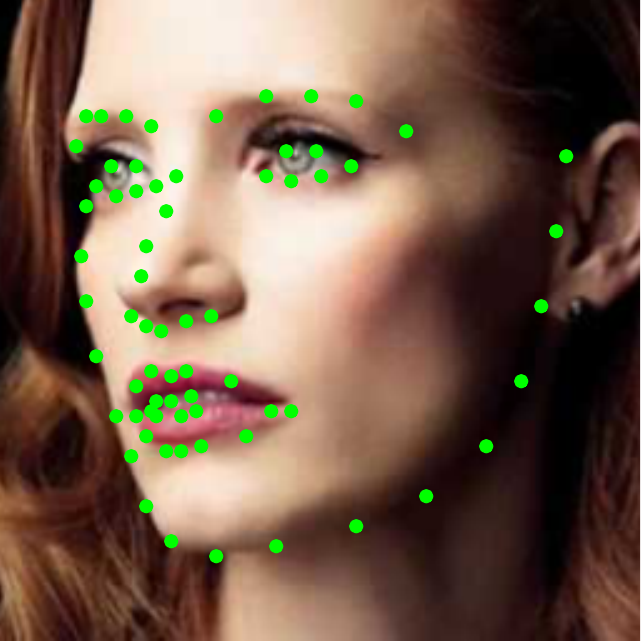}
\includegraphics[width=0.1\textwidth]{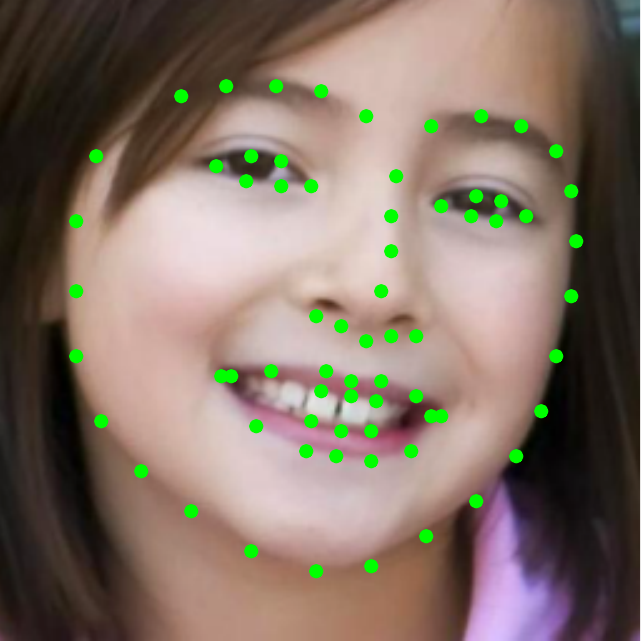}
\vspace{-0.1cm}
\caption{Example alignment results of MobileFAN + KD on $300$W \cite{SagonasTZP13} Full set.}
\label{300w_results}
\vspace{-0.3cm}
\end{figure}

\noindent\textbf{Evaluation on 300W.}
The $300$W dataset is a challenging face alignment benchmark because of its variants on pose and expressions. TABLE \ref{300w_fullset} shows the comparable performance with previous methods on $300$W dataset.
We can observe that simple ``MobileFAN" performs better than the state-of-the-art SAN \cite{dong2018style}, but the number of model parameters of ``MobileFAN" is $28\times$ smaller than that of SAN (we can see form TABLE \ref{runtime}). 
Although ``MobileFAN + KD" does not outperform DCFE \cite{ValleBVB18}, it achieves comparable results to LAB \cite{wu2018look} with extra boundary information on $300$W Full set and Common subset. Using the knowledge distillation, our two full models are better than their corresponding baselines. The ``MobileFAN + KD" achieves $4.17\%$, $4.98\%$ and $3.81\%$ improvements over ``MobileFAN" on $300$W Full set, Challenging subset and Common subset. Although the ``MobileFAN ($0.5$) + KD" fails to compete other state-of-the-art methods, which is possible because the dimension of output score maps (for $300$W dataset, it is $68$) is larger than the dimension of the final deconvolutional layer ($64$), it reduces the mean error from $5.99\%$ to $4.74\%$ on $300$W Full set over its baseline ``MobileFAN ($0.5$)". Fig. \ref{300w_results} visualizes some of our results. It can be observed that driven by knowledge transfer technique, our model can capture various facial expressions accurately.

\begin{table}[t]
\setlength{\tabcolsep}{2.5pt}
\centering
\scriptsize
\caption{Mean error (\%) on $300$W Common subset, Challenging subset and Full set.}
\vspace{-0.3cm}
\begin{tabular}{  l|ccc  }
\hline
Method & Common & Challenging & Full  \\
\hline
RCN \cite{HonariYVP16}& $4.67$ & $8.44$ & $5.41$\\
DAN \cite{KowalskiNT17} & $3.19$ & $5.24$ & $3.59$ \\
PCD-CNN \cite{KumarC18} & $3.67$ & $7.62$ & $4.44$ \\
CPM \cite{dong2018supervision}& $3.39$ & $8.14$ & $4.36$\\
DSRN \cite{miao2018direct}& $4.12$ & $9.68$ & $5.21$\\
SAN \cite{dong2018style}& $3.34$ & $6.60$ & $3.98$ \\
LAB \cite{wu2018look}&2.98 & $5.19 $ & $3.49$ \\  
DCFE \cite{ValleBVB18}& $ 2.76 $ & $5.22$ & $3.24$\\
\hline
Teacher & $2.97$ & $5.23$  & $3.41$\\ 
MobileFAN ($0.5$) & $5.44$ & $8.24$ & $5.99$\\
MobileFAN & $3.10$ & $5.62$ & $3.60$ \\ 
\hline
MobileFAN ($0.5$) + KD & $4.22$ & $6.87$ & $4.74$ \\
MobileFAN + KD & $2.98$  & $5.34$  & $3.45$ \\ 
\hline
\end{tabular}
\label{300w_fullset}
\vspace{-0.6cm}
\end{table}

\noindent\textbf{Evaluation on WFLW.}
A summary of the performance obtained by state-of-the-art methods and the proposed approach on WFLW test set and six subset is shown in TABLE \ref{wflw_testset}. As indicated in TABLE \ref{wflw_testset}, our proposed ``MobileFAN" outperforms LAB \cite{wu2018look} with boundary information in test set and all six subset and ResNet$50$ (Wing + PDB) \cite{FengKA0W18} with strong data augmentation in Test set, Make-up and Occlusion subset. Although ``MobileFAN" performs a little worse than ResNet$50$ (Wing + PDB) \cite{FengKA0W18} in remaining subset, it achieves comparable results with merely $8\%$ of parameters of ResNet$50$ (Wing + PDB) (which can be observed from TABLE \ref{runtime}).
 We can see that ``MobileFAN + KD" model outperforms the-state-of-art methods, with a mean error of $4.93\%$ and failure rate of $5.32\%$ on WFLW test set. In particular, compared with former best models, ResNet$50$ (Wing + PDB) \cite{FengKA0W18} and LAB \cite{wu2018look}, our ``MobileFAN + KD" achieves significant mean error reduction with respect to ResNet$50$ (Wing + PDB) \cite{FengKA0W18} and LAB \cite{wu2018look} of $3.52\%$ and $6.45\%$ on WFLW test set, respectively.
 Similarly, the failure rate is reduced from $6.00\%$ to $5.32\%$ and from $7.56\%$ to $5.32\%$ compared with ResNet$50$ (Wing + PDB) \cite{FengKA0W18} and LAB \cite{wu2018look}.
 
 \begin{table*}[] 
\small
\scriptsize
\centering
\caption{Mean error(\%), failure rate (\%) and AUC on WFLW test set and six subsets: pose, expression (expr.), illumination (illu.), make-up (mu.), occlusion (occu.) and blur.}
\vspace{-0.3cm}
\begin{tabular}{  l|c|c|c|c|c|c|c }
\hline
 Method &test & pose & expr.& illu. & mu. & occu. & blur \\
\hline

\hline
\multicolumn{8}{ l }{Mean Error} \\
\hline
 ESR \cite{CaoWWS12} & $11.13$ & $25.88$ & $11.47$ & $10.49$ & $11.05$ & $13.75$ & $12.20$\\
SDM \cite{XiongT13}& $10.29$ & $24.10$ & $11.45$ & $9.32$ & $9.38$ & $13.03$ & $11.28$\\
CFSS \cite{ZhuLLT15}& $9.07$ & $21.36$ & $10.09$ & $8.30$ & $8.74$ & $11.76$ & $9.96$\\
DVLN \cite{WuY17} & $6.08$ & $11.54$ & $6.78$ & $5.73$ & $5.98$ & $7.33$ & $6.88$\\
LAB  \cite{wu2018look} & $5.27$ & $10.24$ & $5.51$ & $5.23$ & $5.15$ & $6.79$ & $6.32$\\
ResNet$50$(Wing+PDB) \cite{FengKA0W18} & $5.11$& $8.75$ & $5.36$ & $4.93$ & $5.41$ & $6.37$ & $5.81$\\
\hline
Teacher & $4.82$ & $8.48$ & $5.07$ & $4.77$ & $4.55$ & $5.88$ & $5.59$\\
MobileFAN ($0.5$)& $5.70$ &$10.17$ & $5.95$ &$5.68$ &$5.47$ &$6.66$ &$6.49$\\
MobileFAN ($0.5$) + KD & $5.59$ & $9.68$  & $5.98$ & $5.45$ & $5.33$&$6.49$&$6.31$\\
MobileFAN &  $\mathbf{5.09}$ & $8.97$ & $5.43$ & $5.03$ & $\mathbf{4.92}$ & $\mathbf{6.16}$ & $5.95$  \\
MobileFAN + KD & $\mathbf{4.93}$ & $\mathbf{8.72}$ & $\mathbf{5.27}$ & $\mathbf{4.93}$ & $\mathbf{4.70}$&$\mathbf{5.94}$ & $\mathbf{5.73}$\\
\hline

\hline
\multicolumn{8}{ l }{Failure Rate} \\
\hline
 ESR \cite{CaoWWS12} & $35.24$ & $90.18$ & $42.04$ & $30.80$ & $38.84$ & $47.28$ & $41.40$ \\
SDM \cite{XiongT13} & $29.40$ & $84.36$ & $33.44$ &$26.22$& $27.67$ & $41.85$ & $35.32$ \\
CFSS \cite{ZhuLLT15}& $20.56$ & $66.26$ & $23.25$ & $17.34$ & $21.84$ & $32.88$ & $23.67$ \\
DVLN \cite{WuY17}  & $10.84$ & $46.93$ & $11.15$ & $7.31$ & $11.65$ & $16.30$ & $13.71$ \\
LAB \cite{wu2018look} & $7.56$ & $28.83$ & $6.37$ & $6.73$ & $7.77$ & $13.72$ & $10.74$\\
ResNet$50$(Wing+PDB) \cite{FengKA0W18} & $6.00$ & $22.70$ & $4.78$ & $4.30$ & $7.77$ & $12.50$ & $7.76$\\ 
\hline
Teacher& $4.28$ & $20.25$ & $2.55$ & $3.58$ & $3.40$ & $8.56$ & $6.21$\\
MobileFAN ($0.5$)& $7.04$ & $35.89$ & $6.69$ & $6.45$ & $7.77$ & $12.36$ & $7.89$\\
MobileFAN ($0.5$) + KD & $6.72$ & $30.67$ & $6.05$ & $5.73$ & $8.74$ & $12.5$& $8.54$\\
MobileFAN & $\mathbf{5.8}$ & $26.99$ & $\mathbf{4.46}$ & $5.01$ &  $\mathbf{6.80}$ & $\mathbf{11.01}$ & $8.67$\\
MobileFAN + KD & $\mathbf{5.32}$ & $23.93$  & $\mathbf{4.46}$  & $5.01$ &  $\mathbf{6.80}$ & $\mathbf{10.33}$ & $\mathbf{7.24}$\\
\hline

\hline
\multicolumn{8}{ l }{AUC} \\
\hline
ESR \cite{CaoWWS12} & $0.2774$ & $0.0177$ & $0.1981$ & $0.2953$ & $0.2485$ & $0.1946$ & $0.2204$\\
SDM \cite{XiongT13}& $0.3002$ & $0.0226$ & $0.2293$ & $0.3237$ & $0.3125$ & $0.2060$ & $0.2398$\\
CFSS \cite{ZhuLLT15}& $0.3659$ & $0.0632$ & $0.3157$ & $0.3854$ & $0.3691$ & $0.2688$ & $0.3037$\\
DVLN \cite{WuY17}  & $0.4551$ & $0.1474$ & $0.3889$ & $0.4743$ & $0.4494$ & $0.3794$ & $0.3973$\\
LAB \cite{wu2018look} & $0.5323$ & $0.2345$ & $0.4951$ & $0.5433$ & $0.5394$ & $0.4490$ & $0.4630$\\
ResNet$50$(Wing+PDB) \cite{FengKA0W18} & $0.5504$ & $0.3100$ & $0.4959$ & $0.5408$ & $0.5582$ & $0.4885$ & $0.4918$\\
\hline
Teacher &  $0.5414$ & $0.2706$ & $0.5215$ & $0.5513$ & $0.5534$ & $0.4653$ & $0.4775$\\
MobileFAN ($0.5$)& $0.4630$ & $0.1605$ & $0.4380$ & $0.4758$ & $0.4777$ & $0.4029$ &$0.4065$\\
MobileFAN ($0.5$) + KD &$0.4682$ &$0.1728$ & $0.4373$& $0.4819$ & $0.4823$&$0.4028$&$0.4091$\\
MobileFAN &$0.5163$ & $0.2270$ & $0.4888$ &$0.5276$ & $0.5251$ &$0.4417$ &$0.4484$ \\
MobileFAN + KD & $0.5296$ & $0.2443$  & $0.5039$ & $0.5388$  & $0.5442$ & $0.4576$ & $0.4661$ \\
\hline
\end{tabular}
\label{wflw_testset}
\end{table*}
 
\begin{figure}[t]
\centering
\includegraphics[width=8cm]{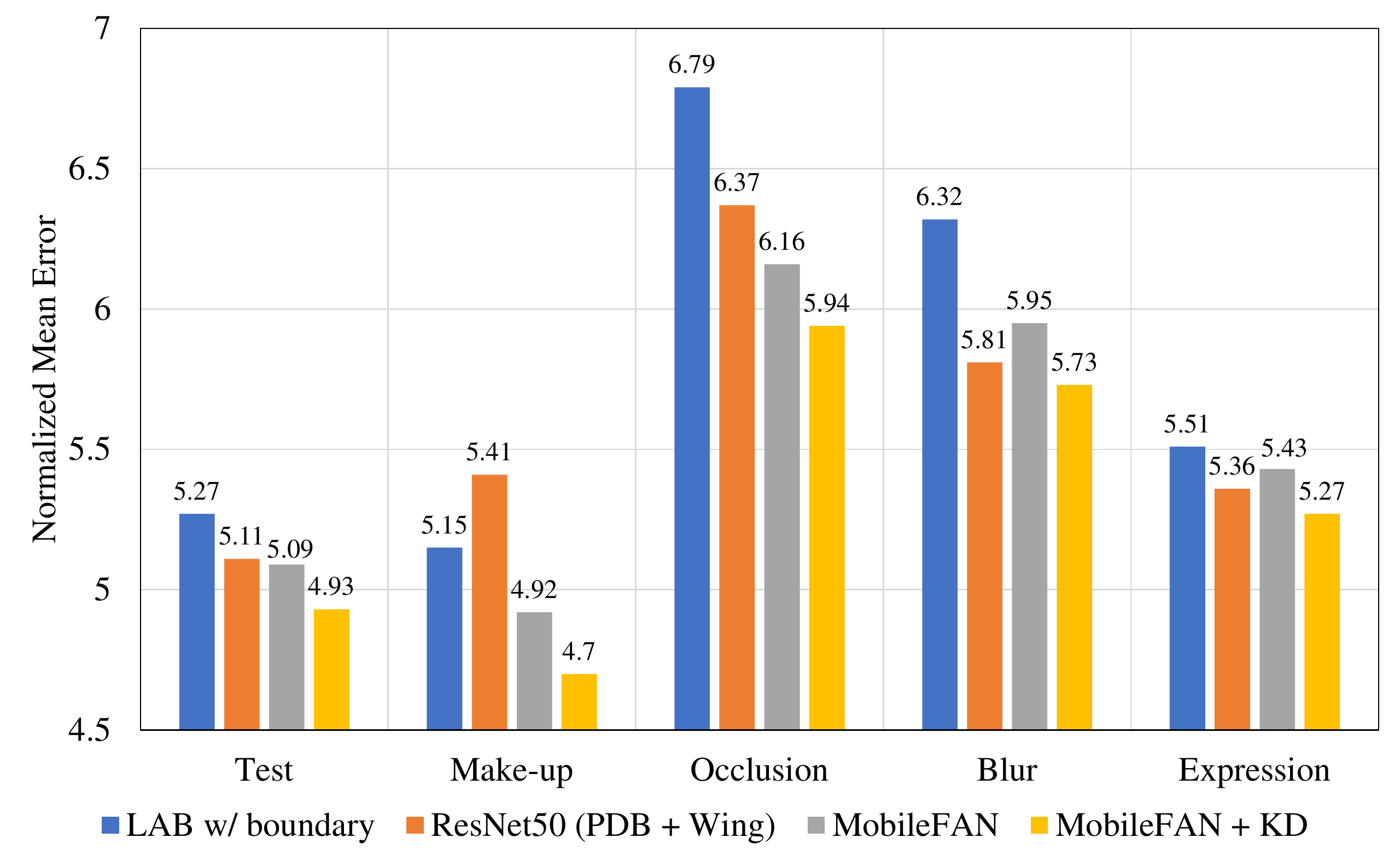}
\caption{Normalized Mean Error (\%) on WFLW \cite{wu2018look} Test set and $4$ typical subset for the method of LAB \cite{wu2018look}, ResNet$50$ (Wing + PDB) \cite{FengKA0W18} and the proposed method.}
\label{wflw_bar}
\vspace{-0.6cm}
\end{figure}

 \begin{figure}[t]
\centering
\subfigure[0.08]{
\begin{minipage}[t]{0.18\linewidth}
\centering
\includegraphics[width=0.7in]{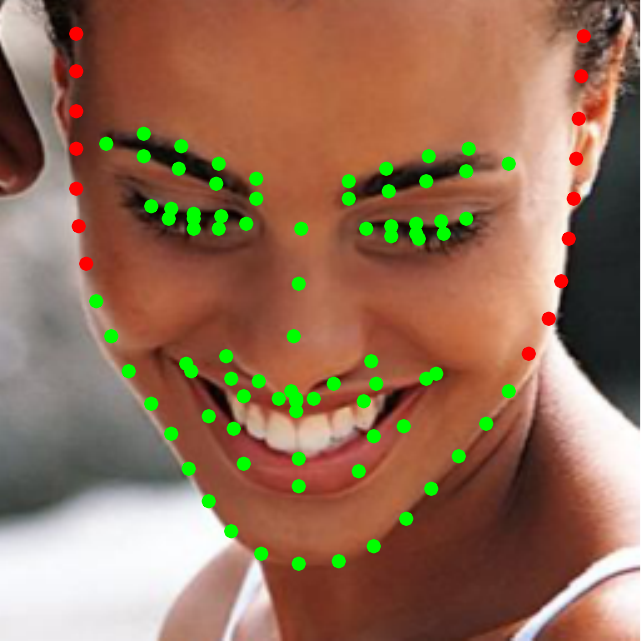}
\includegraphics[width=0.7in]{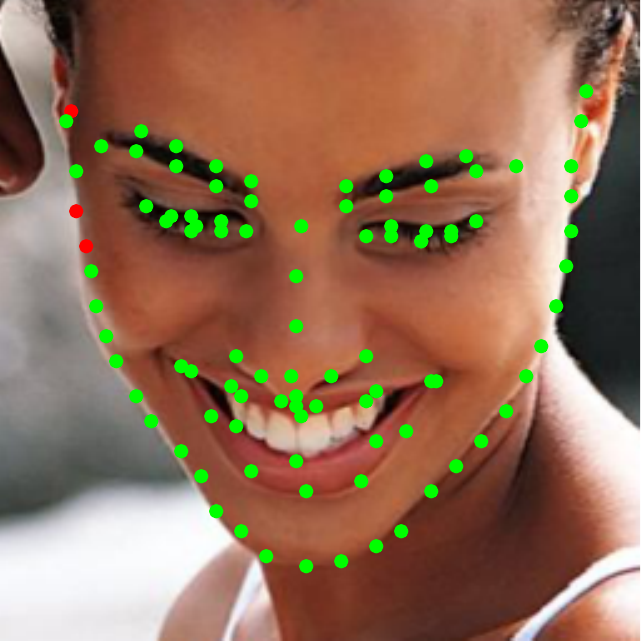}
\includegraphics[width=0.7in]{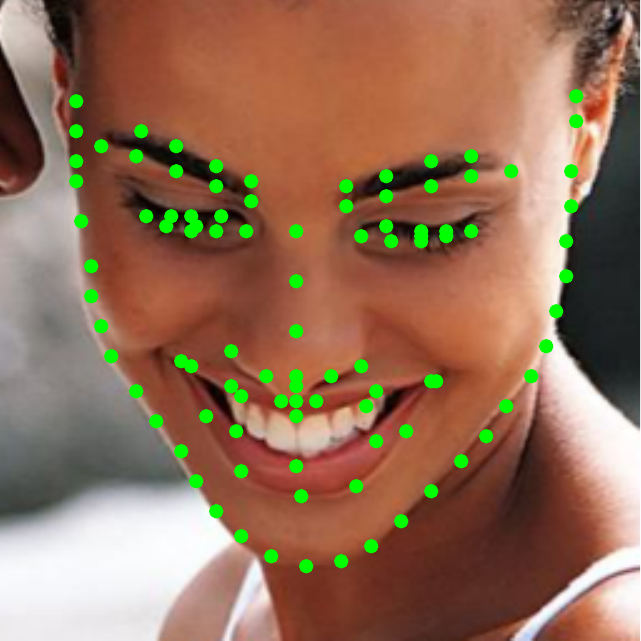}
\end{minipage}
}
\subfigure[0.04]{
\begin{minipage}[t]{0.18\linewidth}
\centering
\includegraphics[width=0.7in]{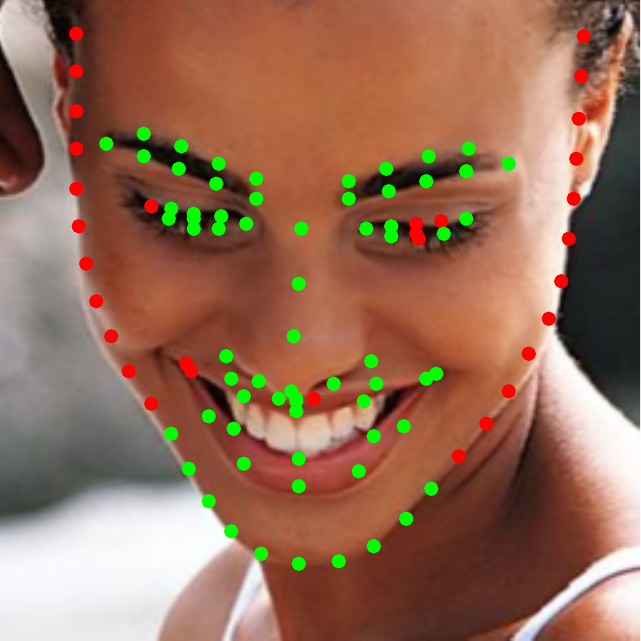}
\includegraphics[width=0.7in]{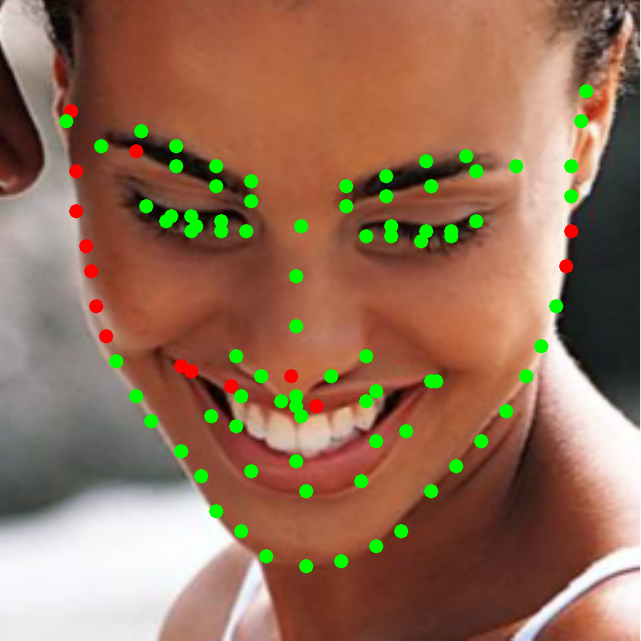}
\includegraphics[width=0.7in]{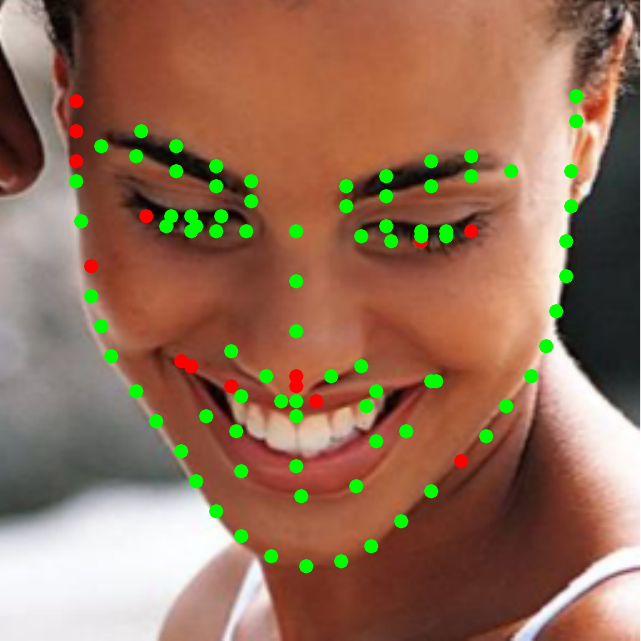}
\end{minipage}
}
\subfigure[0.08]{
\begin{minipage}[t]{0.18\linewidth}
\centering
\includegraphics[width=0.7in]{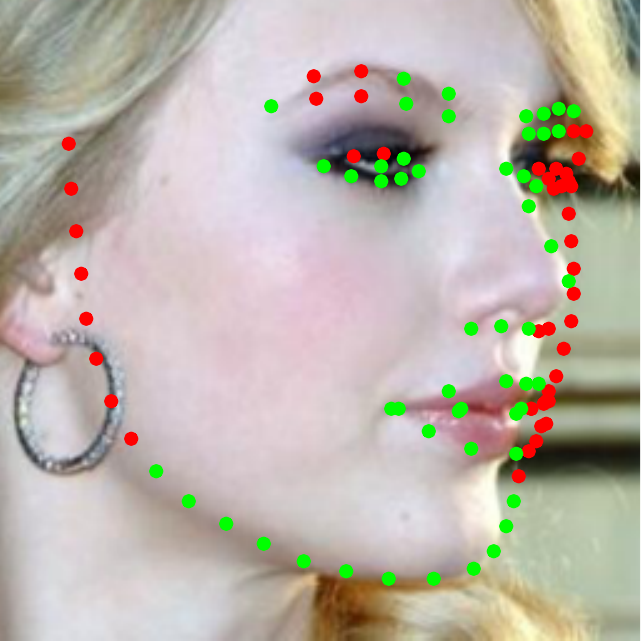}
\includegraphics[width=0.7in]{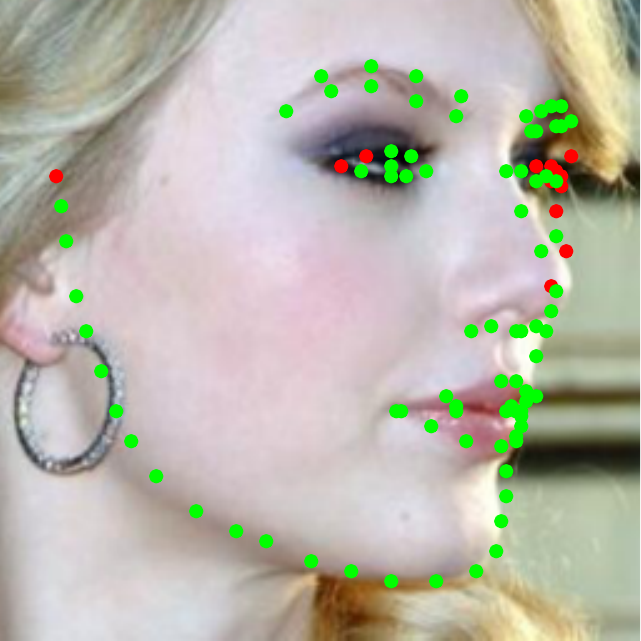}
\includegraphics[width=0.7in]{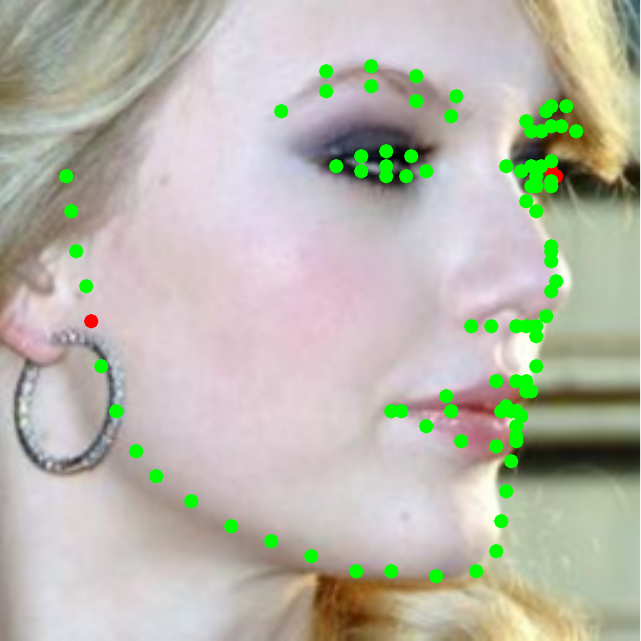}
\end{minipage}
}
\subfigure[0.04]{
\begin{minipage}[t]{0.18\linewidth}
\centering
\includegraphics[width=0.7in]{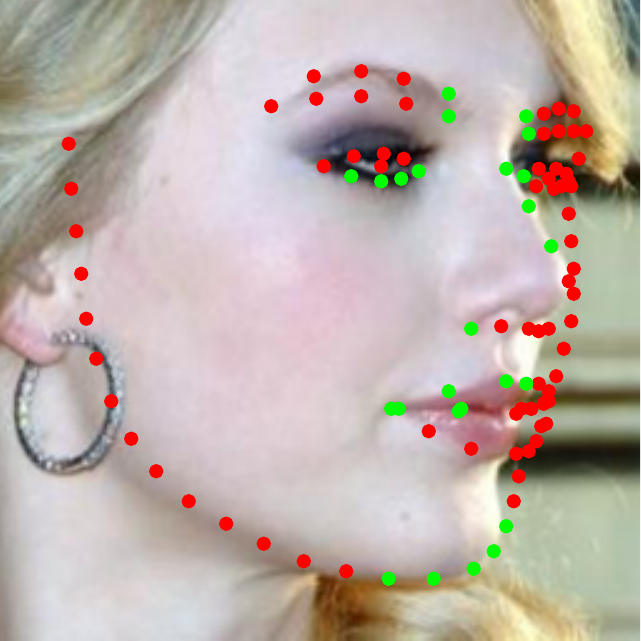}
\includegraphics[width=0.7in]{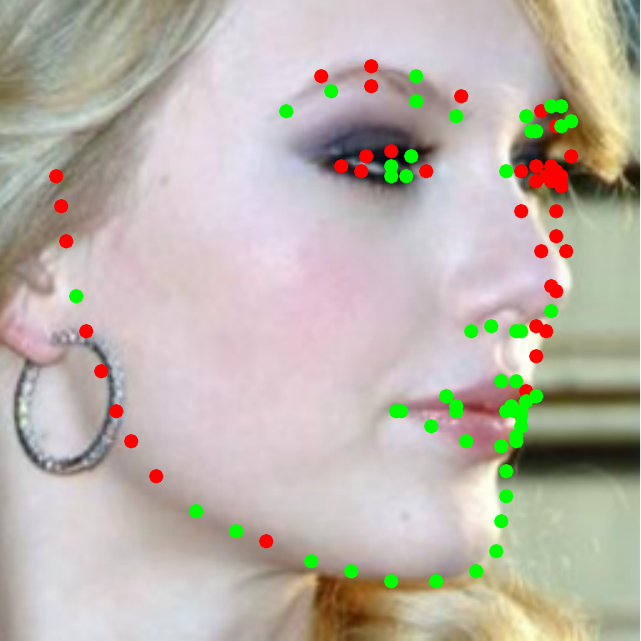}
\includegraphics[width=0.7in]{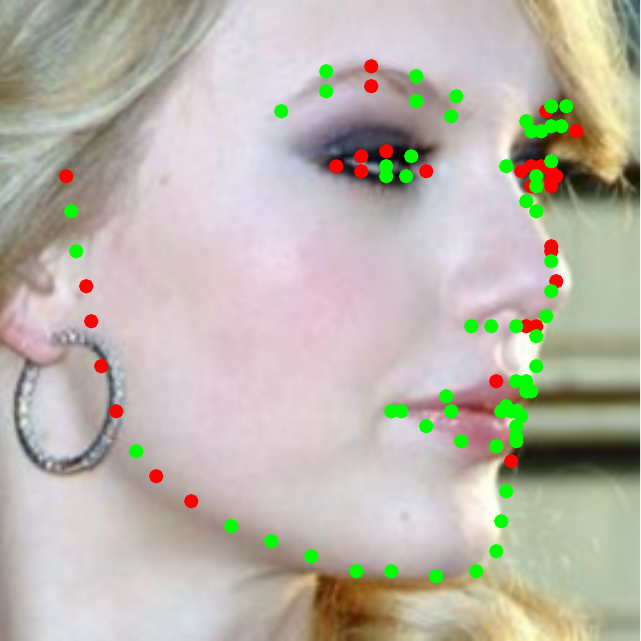}
\end{minipage}
} 
\centering
\vspace{-0.3cm}
\caption{
Results of example samples of WFLW \cite{wu2018look} under different mean error thresholds: 0.08 and 0.04. Top row: Results of LAB \cite{wu2018look}. Middle row: Results of MobileFAN. Bottom row: Results of MobileFAN + KD. Red points indicate normalized mean error is lager than threshold.
}
\label{wflw_parts}
\vspace{-0.8cm}
\end{figure}

\begin{figure*}[ht] 
\centering
\includegraphics[width=0.1\textwidth]{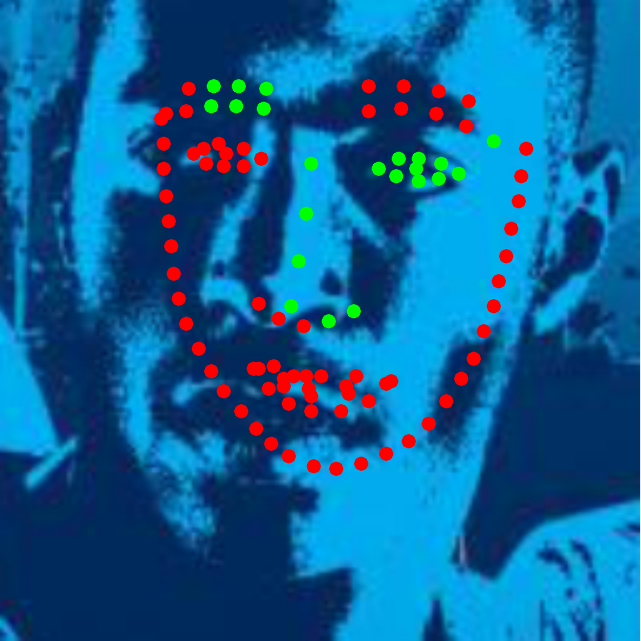}
\includegraphics[width=0.1\textwidth]{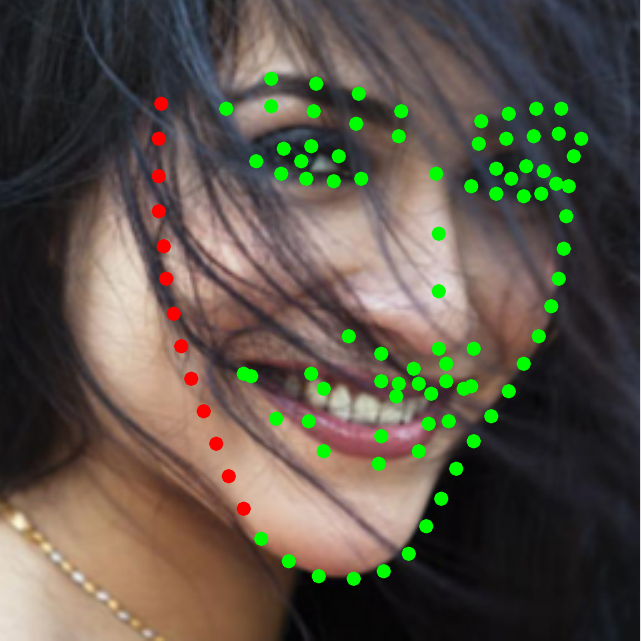}
\includegraphics[width=0.1\textwidth]{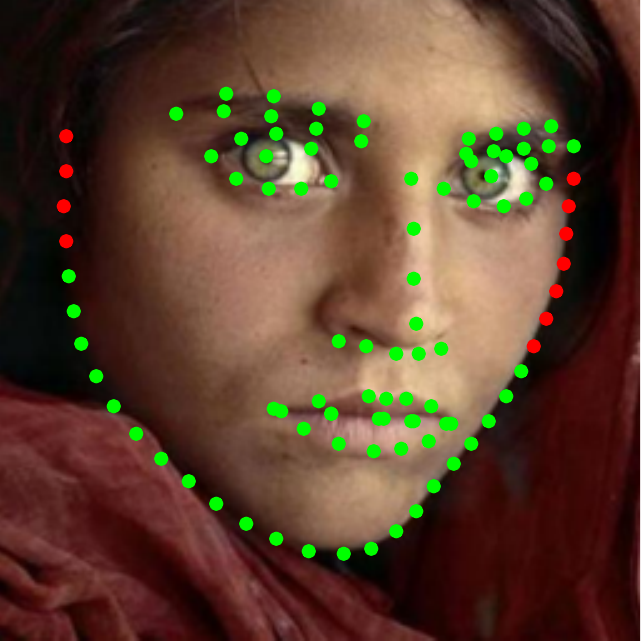}
\includegraphics[width=0.1\textwidth]{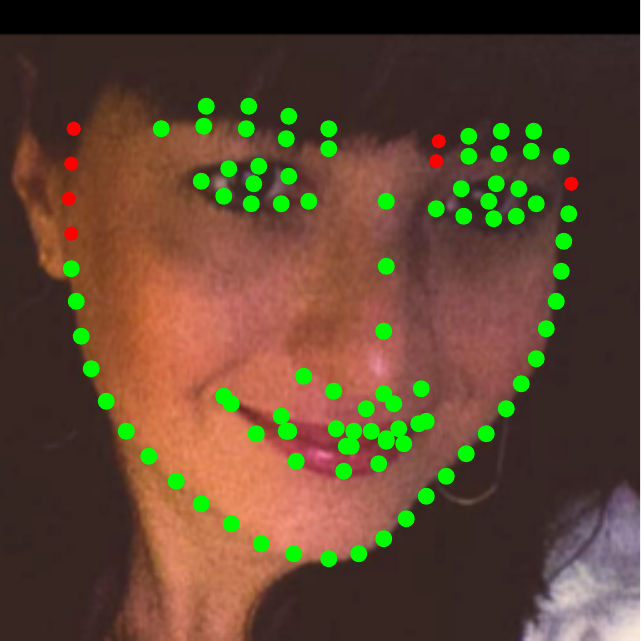}
\includegraphics[width=0.1\textwidth]{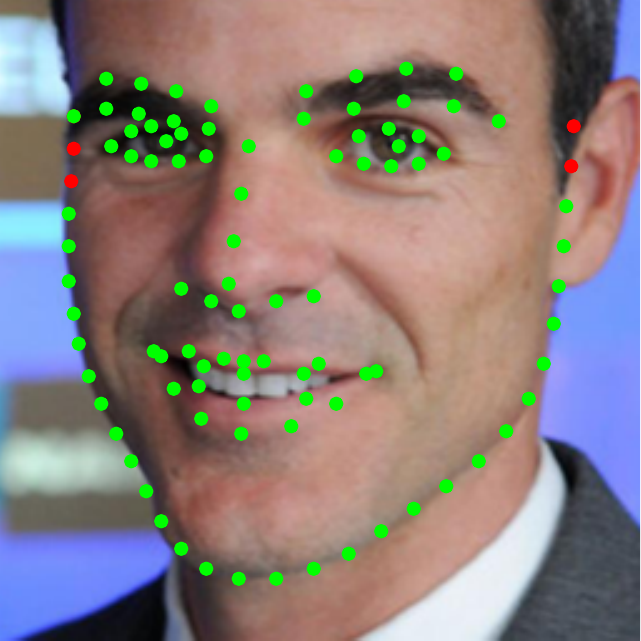}
\includegraphics[width=0.1\textwidth]{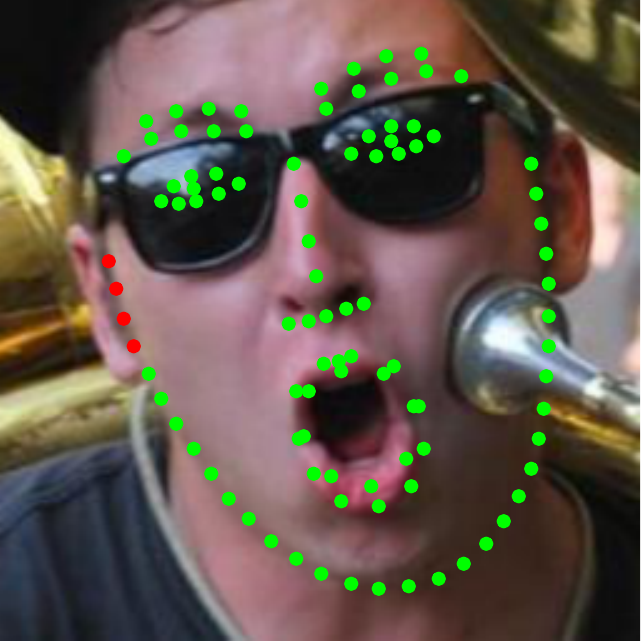}
\includegraphics[width=0.1\textwidth]{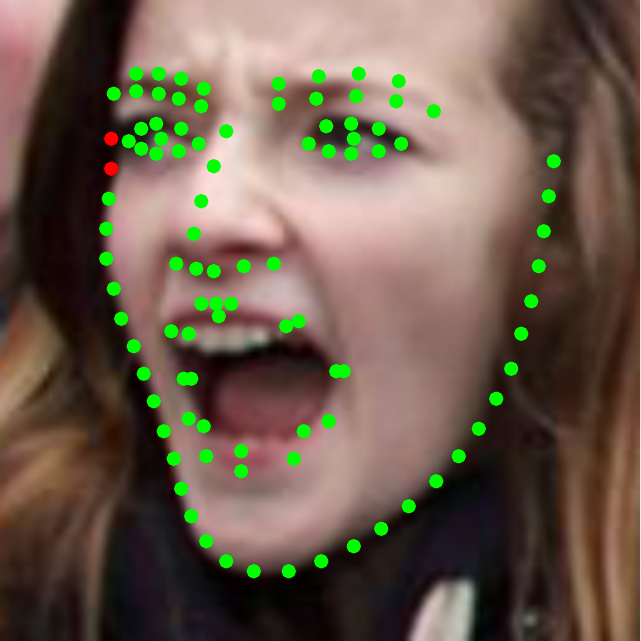}
\includegraphics[width=0.1\textwidth]{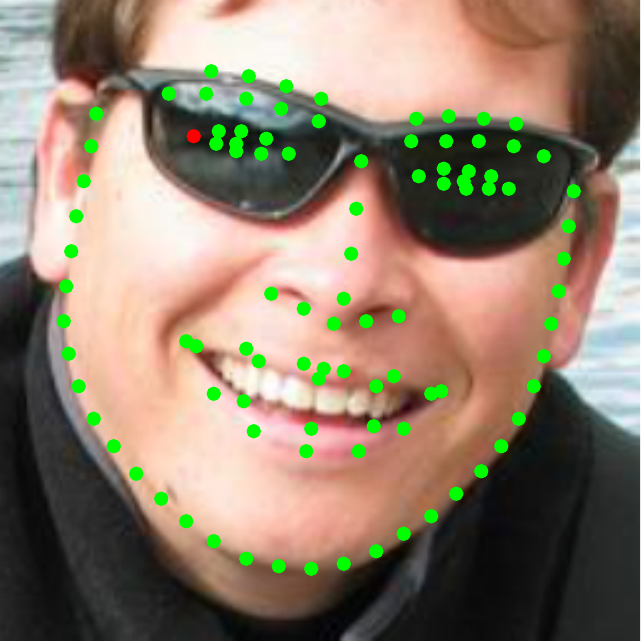}
\includegraphics[width=0.1\textwidth]{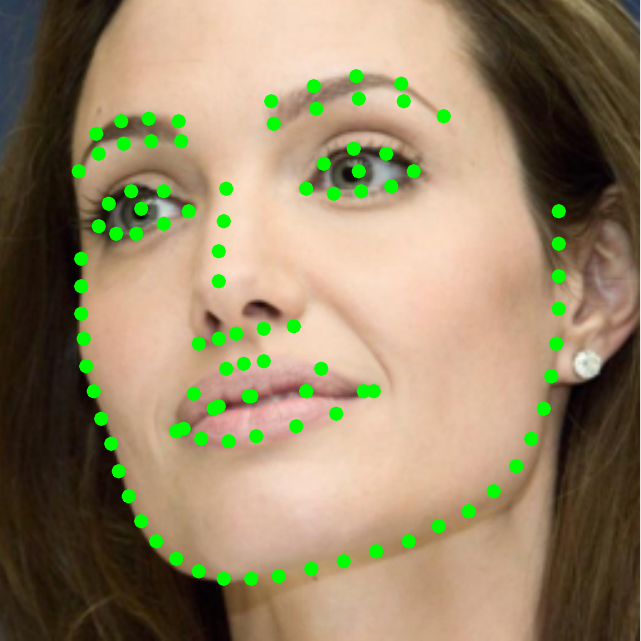}

\includegraphics[width=0.1\textwidth]{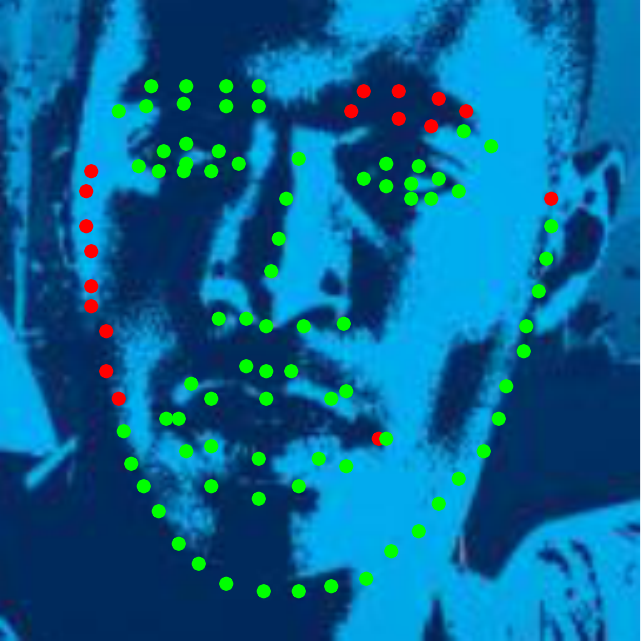}
\includegraphics[width=0.1\textwidth]{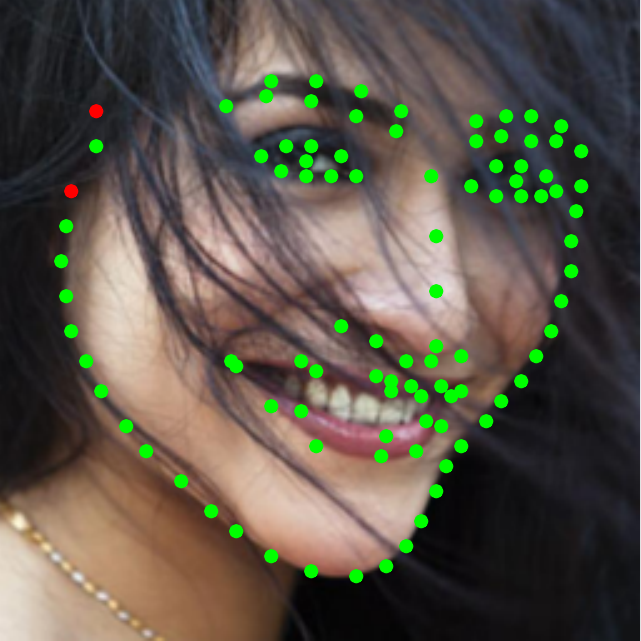}
\includegraphics[width=0.1\textwidth]{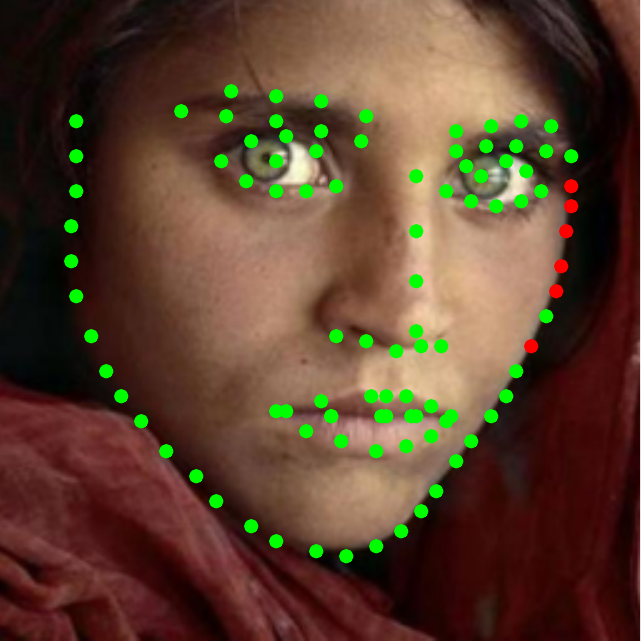}
\includegraphics[width=0.1\textwidth]{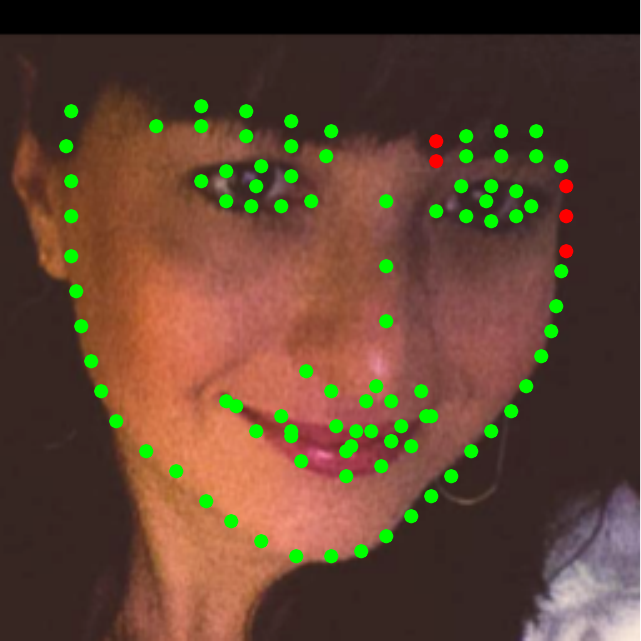}
\includegraphics[width=0.1\textwidth]{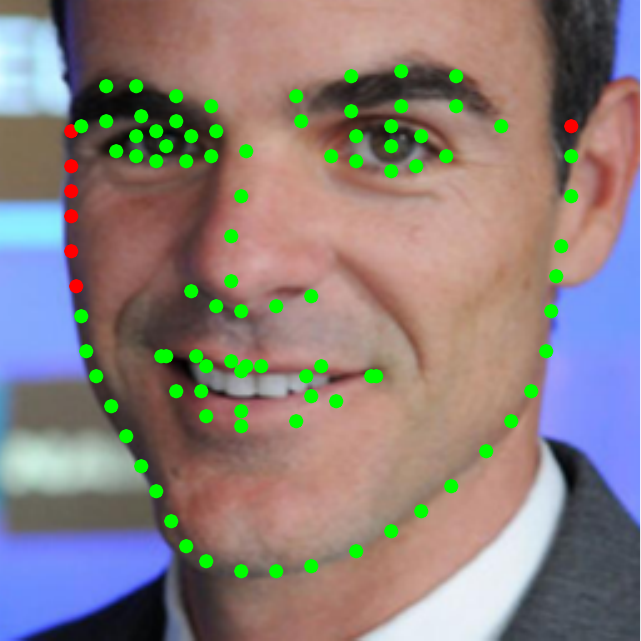}
\includegraphics[width=0.1\textwidth]{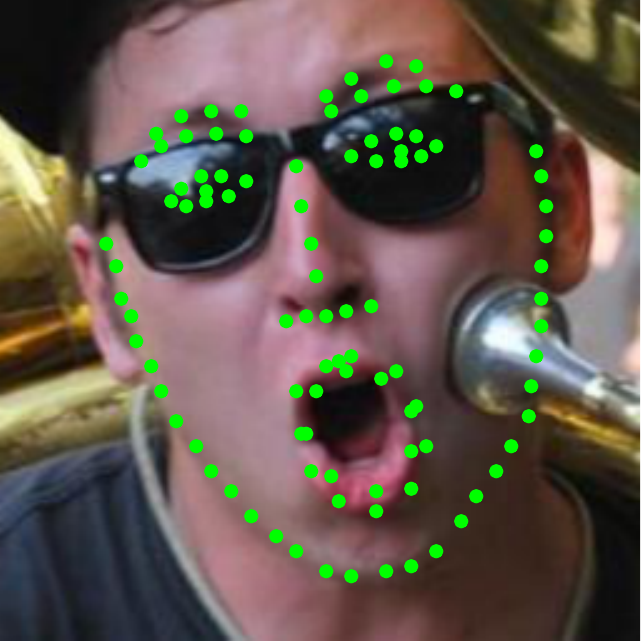}
\includegraphics[width=0.1\textwidth]{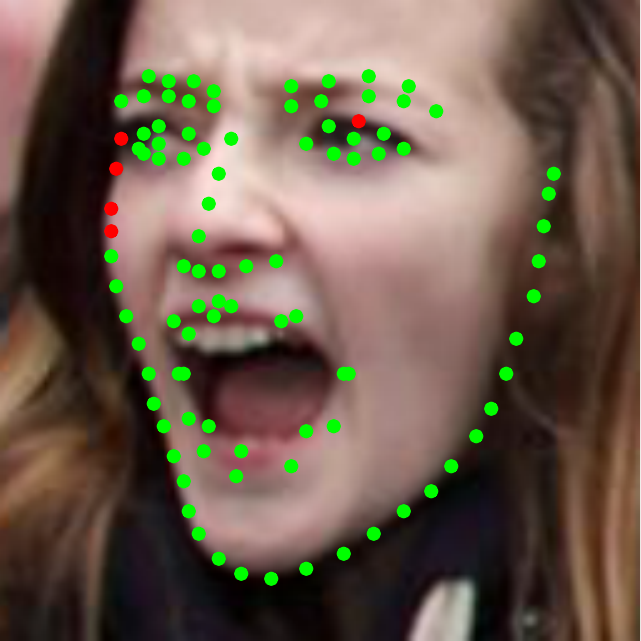}
\includegraphics[width=0.1\textwidth]{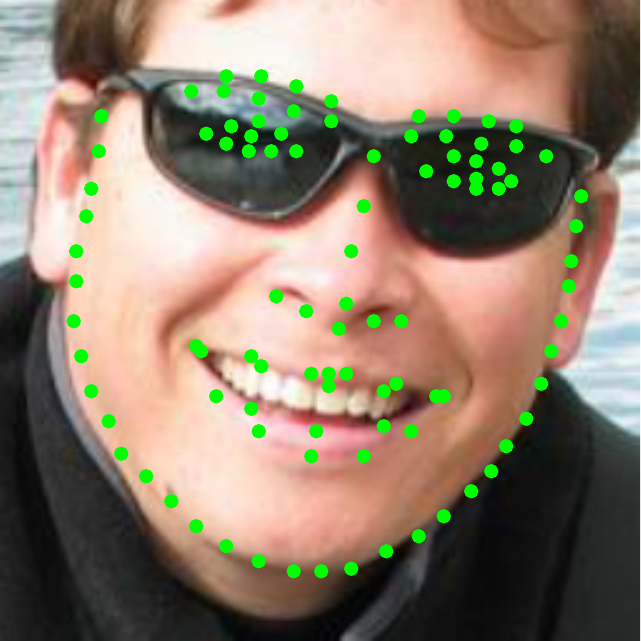}
\includegraphics[width=0.1\textwidth]{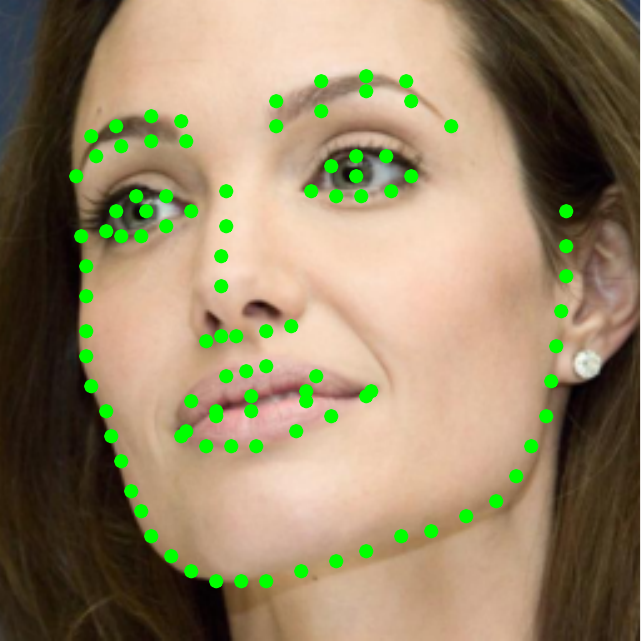}

\includegraphics[width=0.1\textwidth]{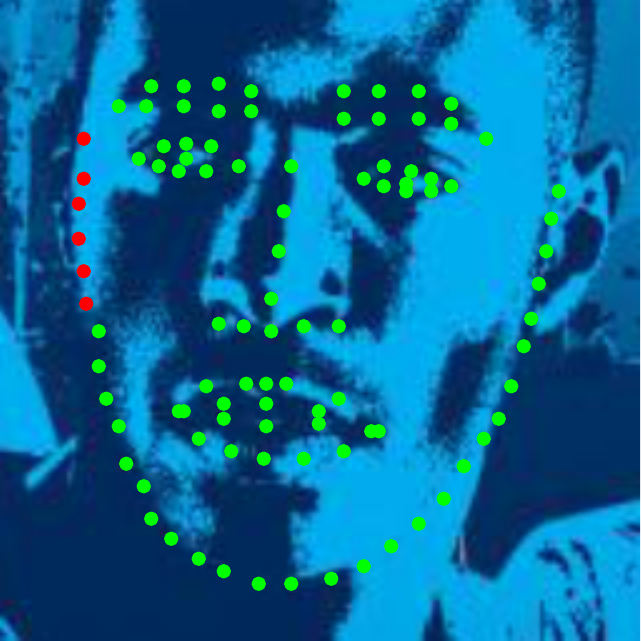}
\includegraphics[width=0.1\textwidth]{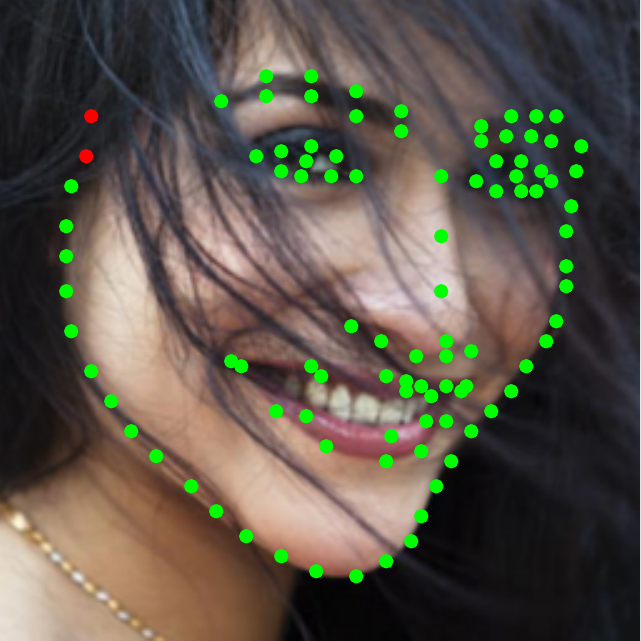}
\includegraphics[width=0.1\textwidth]{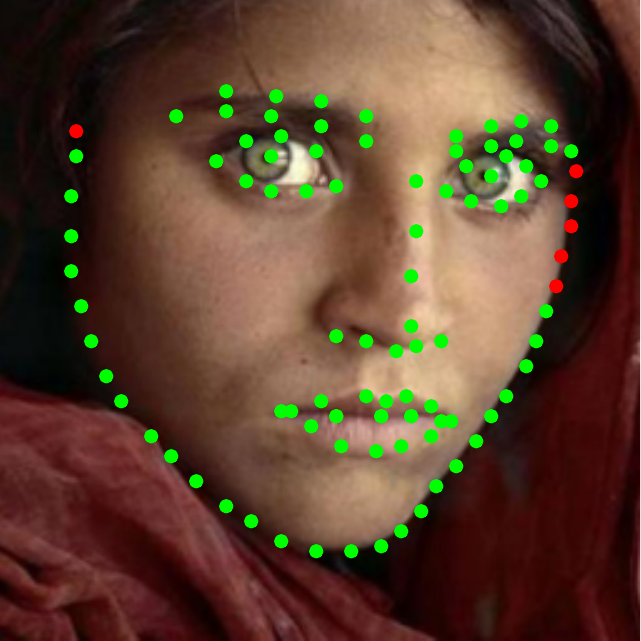}
\includegraphics[width=0.1\textwidth]{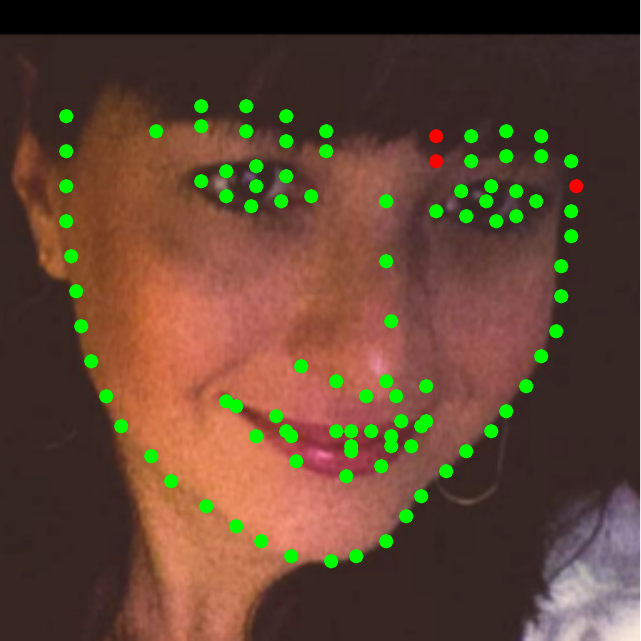}
\includegraphics[width=0.1\textwidth]{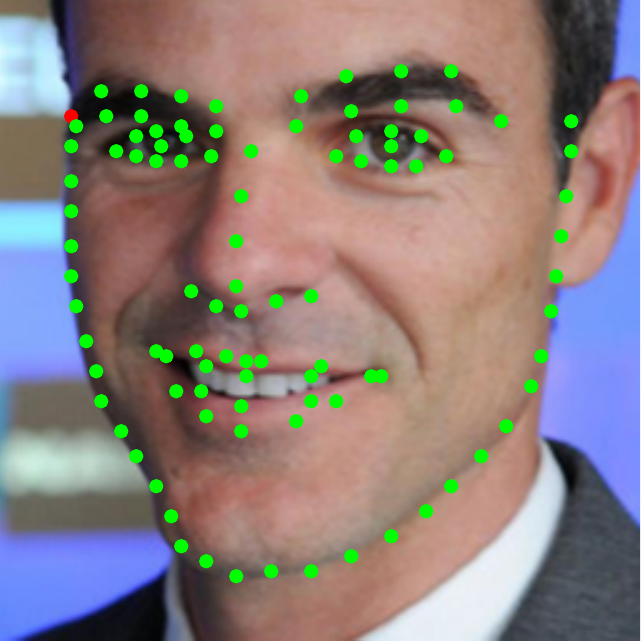}
\includegraphics[width=0.1\textwidth]{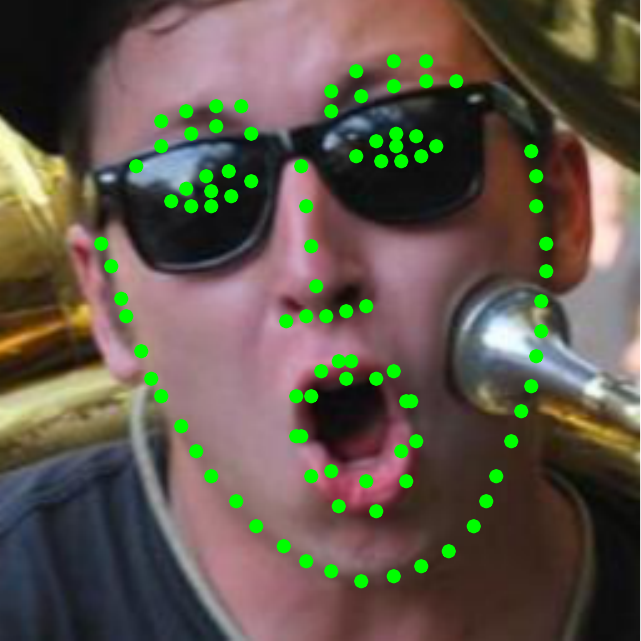}
\includegraphics[width=0.1\textwidth]{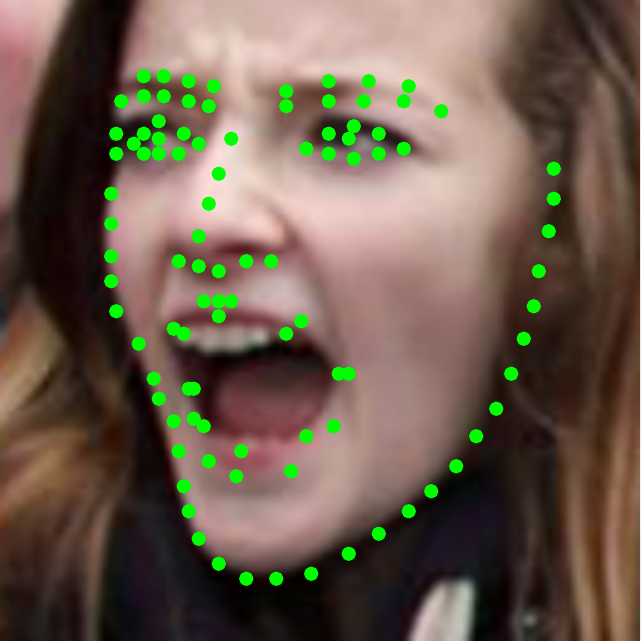}
\includegraphics[width=0.1\textwidth]{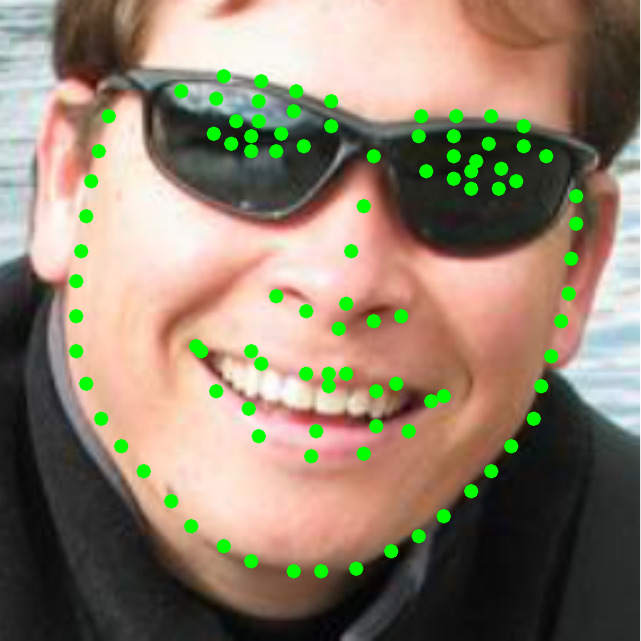}
\includegraphics[width=0.1\textwidth]{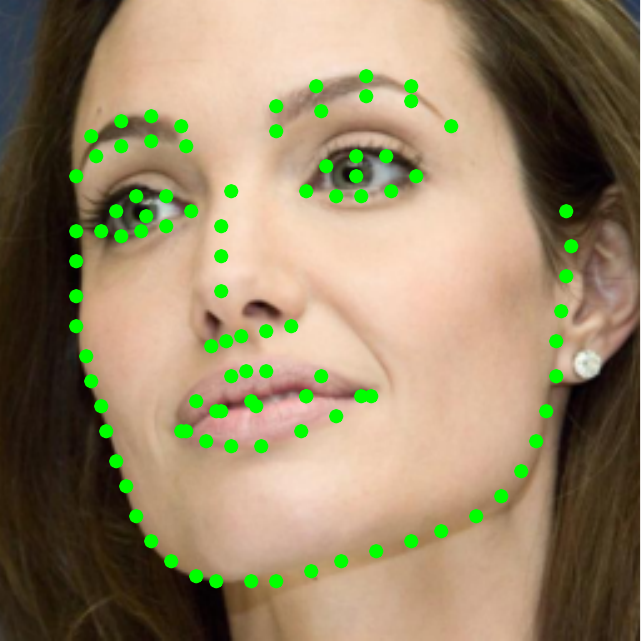}

\includegraphics[width=0.1\textwidth]{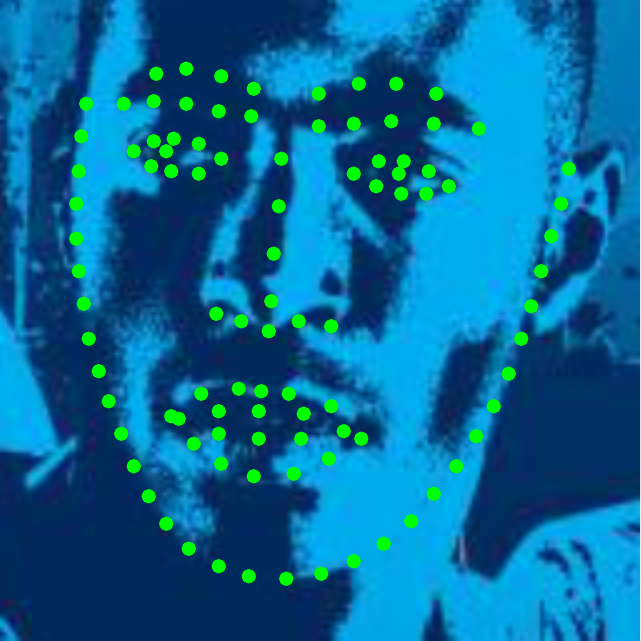}
\includegraphics[width=0.1\textwidth]{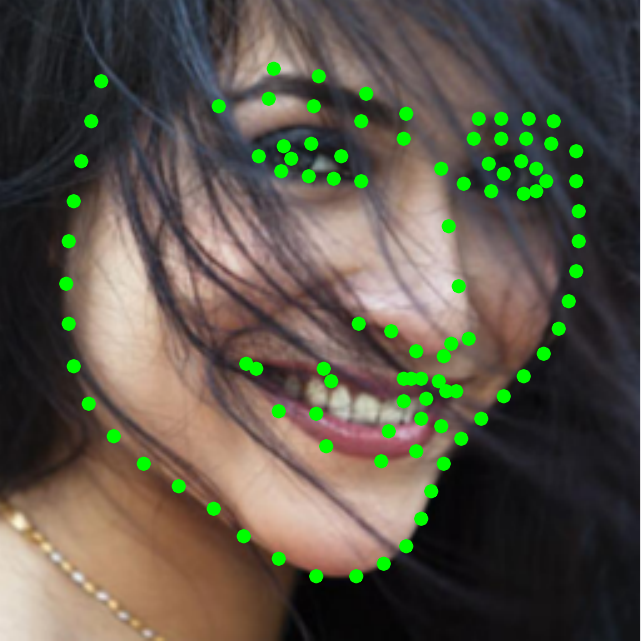}
\includegraphics[width=0.1\textwidth]{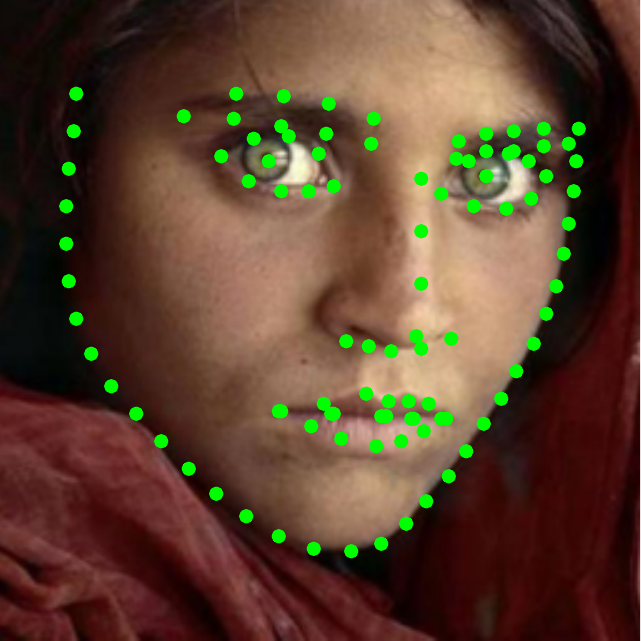}
\includegraphics[width=0.1\textwidth]{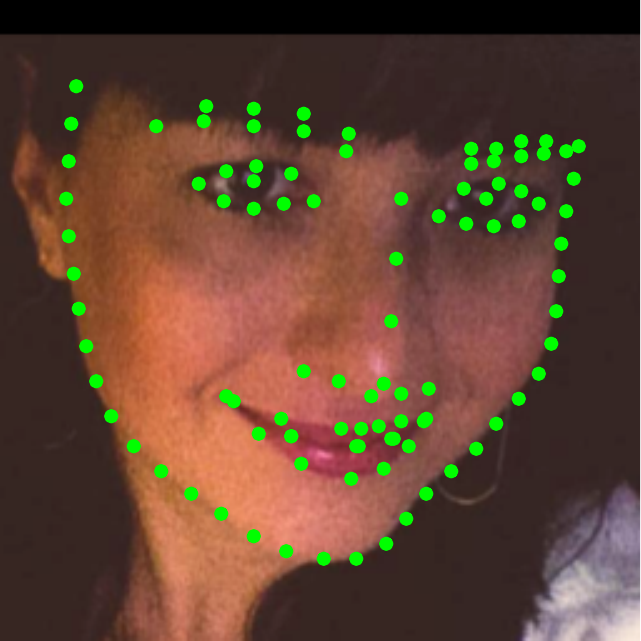}
\includegraphics[width=0.1\textwidth]{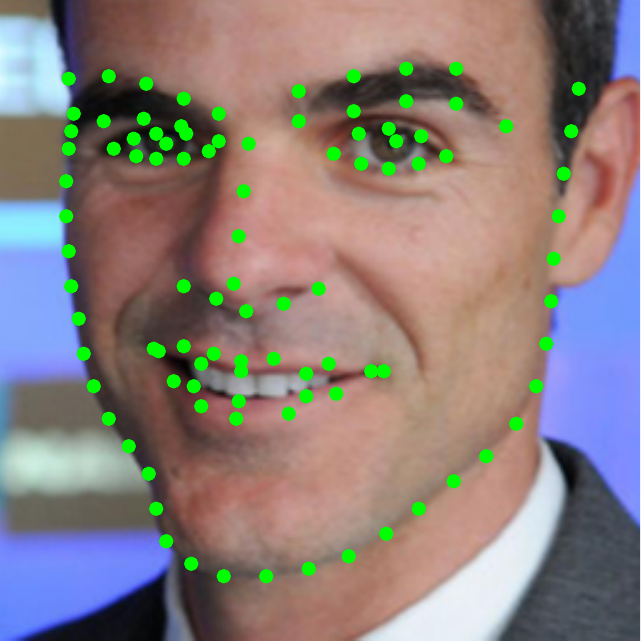}
\includegraphics[width=0.1\textwidth]{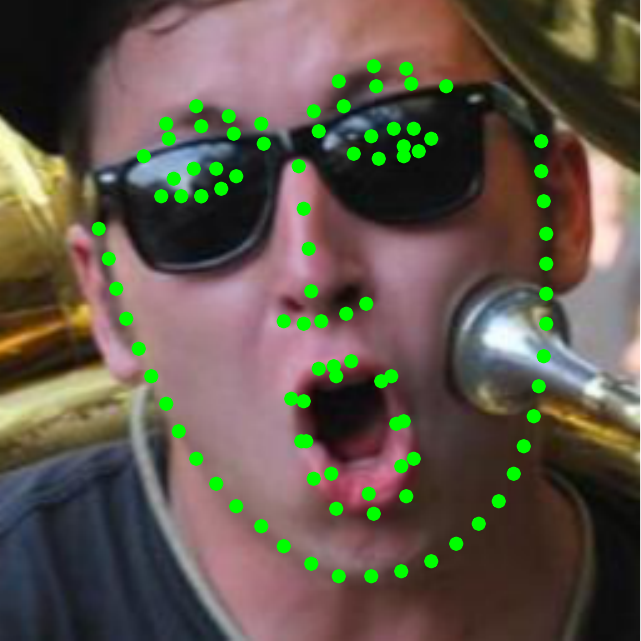}
\includegraphics[width=0.1\textwidth]{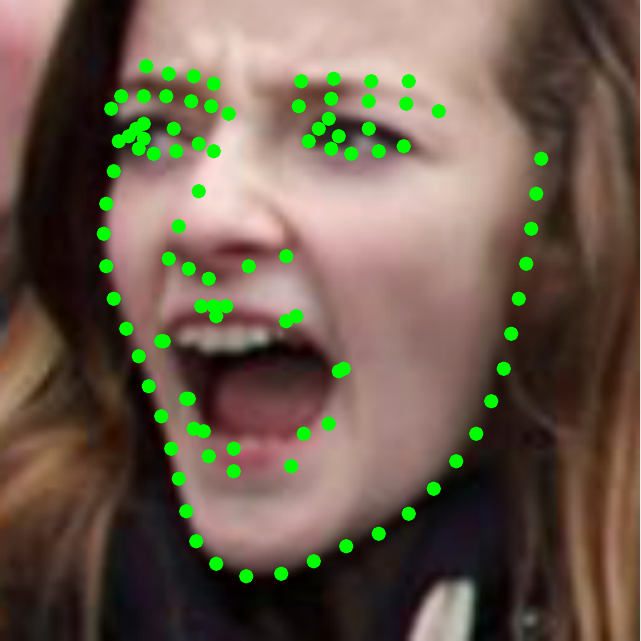}
\includegraphics[width=0.1\textwidth]{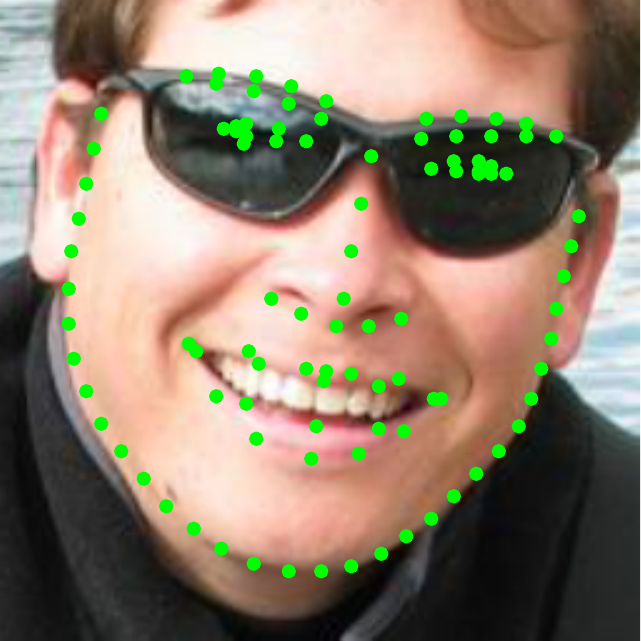}
\includegraphics[width=0.1\textwidth]{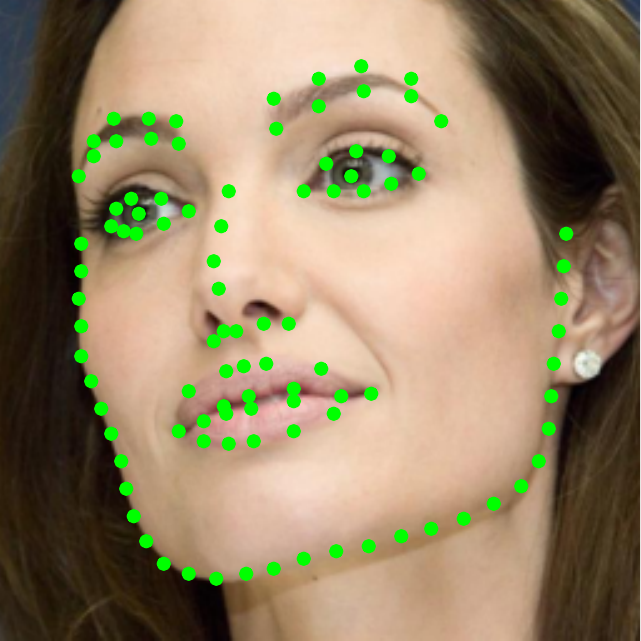}

\caption{Example face alignment results on WFLW \cite{wu2018look} test set. Top row: Face alignment results of LAB \cite{wu2018look}. Second row: Face alignment results of our proposed MobileFAN. Third row: Face alignment results of our MobileFAN + KD. Bottom row: Ground-truth annotations. Red points indicate normalized error is larger than $0.1$.}
\label{wflw_results}
\vspace{-0.7cm}
\end{figure*}
 
 To provide a more straightforward comparative illustration, we compare in Fig. \ref{wflw_bar} our ``MobileFAN" and ``MobileFAN + KD" against ResNet$50$ (Wing + PDB) and LAB on WFLW test set and four typical subsets. 
 We can see that ``MobileFAN" outperforms ResNet$50$ (Wing + PDB) and LAB with a big margin on WFLW Test set, Make-up subset and Occlusion subset, let along ``MobileFAN + KD" with the help of the ``Teacher".
 In particular, ``MobileFAN" achieves $9.06\%$ relative improvement in mean error reduction over ResNet$50$ (Wing + PDB) on Make-up subset, as well as $9.28\%$ relative improvement in mean error reduction over LAB on Occlusion subset.
 Although ``MobileFAN" achieves comparable results compared with ResNet$50$ (Wing + PDB) on Blur subset and Expression subset, it outperforms LAB with a big margin by using less number of parameters. With the help of knowledge distillation, ``MobileFAN + KD" achieves the state-of-the-art performance on WFLW test set and six subsets.
The results indicate that our proposed lightweight model is robust to extreme conditions.
We can visually see the advantages of our ``MobileFAN" from Fig. \ref{wflw_parts}. Specifically, we compare LAB, ``MobileFAN" and ``MobileFAN + KD" under different mean error thresholds. It can be found that the number of landmarks of low mean error of our method is more than that of LAB. And the third row in Fig. \ref{wflw_parts} depicts further improvements led by adding feature-aligned distillation and feature-similarity distillation, where the knowledge transfer techniques provide richer facial details to make relative spatial positions between facial landmarks more precisely.

 Some example results of LAB\footnote{Model available from https://github.com/wywu/LAB.}, ``MobileFAN", ``MobileFAN +KD" and ground-truth on WFLW test set is showed in Fig. \ref{wflw_results}. We can observe that ``MobileFAN + KD" improves the accuracy of landmarks above the face contour (chin), eyebrow, eye corner and so on.

 \noindent\textbf{Model Size and Computational Cost Analysis.}
 To further evaluate the model size and the computational complexity, we calculate the number of network parameters (\#Params), the sum of floating point operations (FLOPs) and the speed of our approach and other competing methods. The FLOPs of our model is calculated on the resolution of $256 \times 256$.
Frames per second (fps) is adopted for measuring the computation speed.
Here the fps calculation is performed on an NVIDIA GeForce GTX 1070 card.
We notice that the model size of our compact network is the smallest.
We can see from TABLE \ref{runtime} that the proposed models have minimal parameters and lowest computation complexity against LAB \cite{wu2018look} and ResNet$50$ (Wing + PDB) \cite{FengKA0W18}, while remaining effective for facial landmark localization. Specifically, MobileFAN ($0.5$) and MobileFAN just have $1.84$M and $2.02$M parameters, respectively. Although our MobileFAN has fewer parameters, \textit{e.g.}, $8\%$ of model size of LAB and ResNet$50$ (Wing + PDB) and $3.54\%$ of model size of SAN, it achieves comparable or even better results competing with the state-of-the-art methods.
It is observed that the proposed MobileFAN can process 238 fps, which is significantly faster than the 6, 50, and 126 fps of
the state-of-the-art approaches.

\begin{table}[ht]
\setlength{\tabcolsep}{2.5pt}
\centering
\scriptsize
\vspace{-0.3cm}
\caption{A comparison of different networks in backbone, model size (the number of
model parameters), computational cost (FLOPS) and speed (fps). }
\vspace{-0.2cm}
\begin{tabular}{  l|cccc  }
\hline
Method & backbone & \#Params (M) & FLOPs (B) & Speed(fps) \\
\hline
DVLN \cite{WuY17} & VGG-$16$ & $132.0$ & $14.4$ & -\\
SAN \cite{dong2018style} & ResNet-$152$& $57.4$  & $10.7$ & $50$\\
LAB \cite{wu2018look} & Hourglass &$25.1$ &$19.1$ & $6$\\
ResNet$50$ (Wing + PDB) \cite{FengKA0W18} & ResNet-$50$ & $25$ & $3.8$ & $126$ \\
\hline
MobileFAN & MobileNetV$2$ &$2.02$ &$0.72$ & $238$ \\
MobileFAN ($0.5$) & MobileNetV$2$ & $\mathbf{1.84}$ &$\mathbf{0.45}$ & $\mathbf{249}$ \\
\hline
\end{tabular}
\label{runtime}
\vspace{-0.7cm}
\end{table}

\subsection{Ablation study}
Our framework consists of several different components, such as feature-aligned distillation and feature-similarity distillation of different deconvolutional layers. In this section, we take a look at the effectiveness of different distillation methods on $300$W Challenging subset.
Based on the baseline network, MobileFAN without distillation (``MobileFAN"), we evaluate the mean error using various combinations of each component, as summarized in TABLE \ref{component}.
In addition, we analyze the influence of the hyperparameter $\lambda$ (as described in Section \uppercase\expandafter{\romannumeral3}) on COFW Test set.

\begin{table}[t]
\setlength{\tabcolsep}{5pt}
\scriptsize
\centering
\caption{The proposed distillation components in our method.}
\vspace{-0.3cm}
\begin{tabular}{  l|c }
\hline
Proposed distillation component & Abbreviation\\
\hline
feature-aligned distillation of the first deconvolutional layer & FA$_1$ \\
feature-aligned distillation of the second deconvolutional layer & FA$_2$ \\
feature-aligned distillation of the third deconvolutional layer & FA$_3$ \\
\hline
feature-similarity distillation of the first deconvolutional layer & FS$_1$ \\
feature-similarity distillation of the second deconvolutional layer & FS$_2$ \\
feature-similarity distillation of the third deconvolutional layer & FS$_3$ \\
\hline
\end{tabular}
\label{component}
\vspace{-0.3cm}
\end{table}

\vspace{-0.2cm}
\noindent\textbf{Feature-aligned distillation over layer.}
To investigate the effectiveness of the feature-aligned distillation in facial landmark detection, we implement the feature-aligned distillation by adopting a $1 \times 1$ convolution layer to align the feature of each pixel between teacher network $\mathcal{T}$ and student network $\mathcal{S}$, such that the channel of the features is matched.

\begin{figure}[ht] 
\centering
\includegraphics[width=0.1\textwidth]{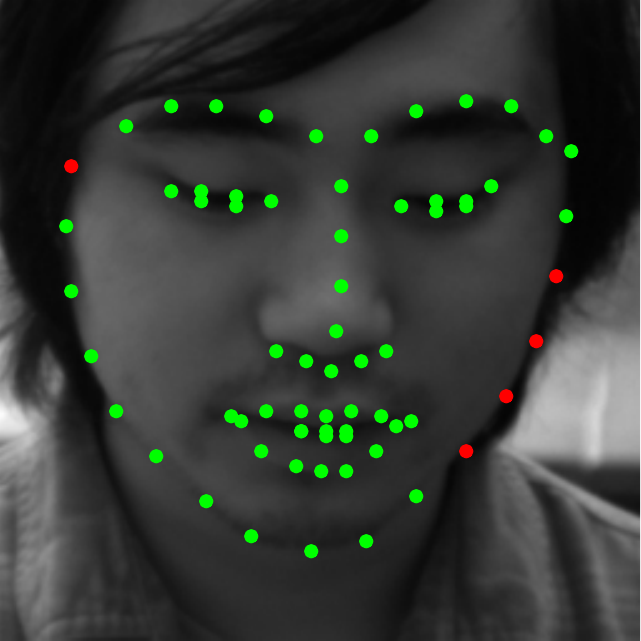}
\includegraphics[width=0.1\textwidth]{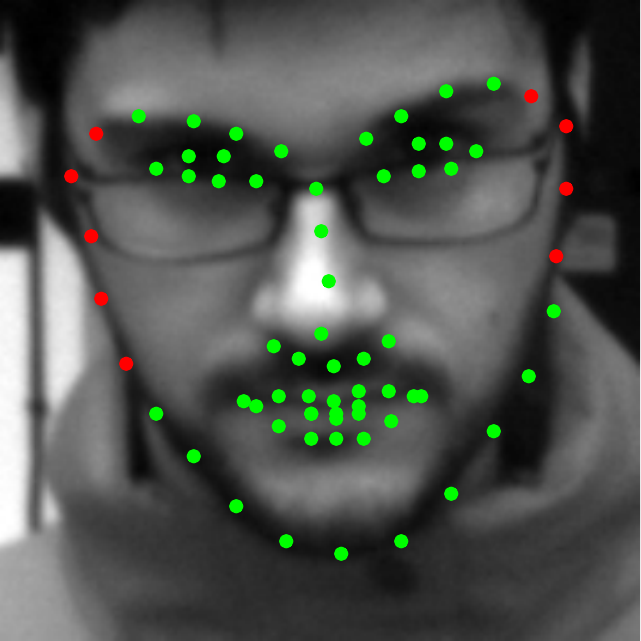}
\includegraphics[width=0.1\textwidth]{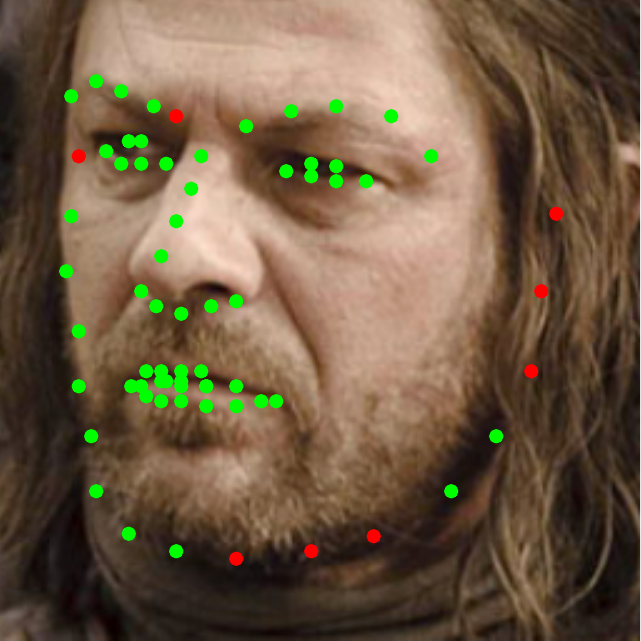}
\includegraphics[width=0.1\textwidth]{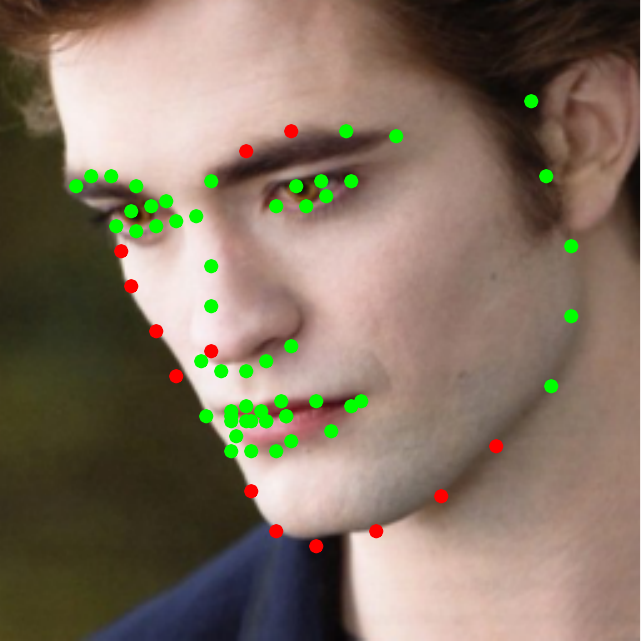}

\includegraphics[width=0.1\textwidth]{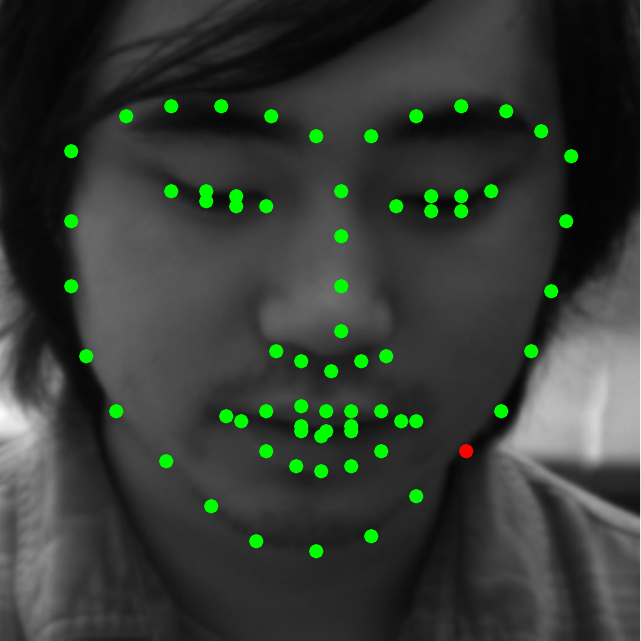}
\includegraphics[width=0.1\textwidth]{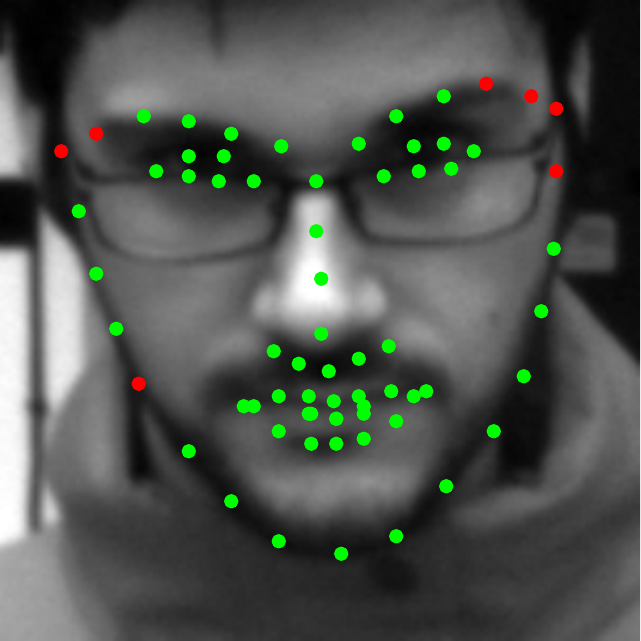}
\includegraphics[width=0.1\textwidth]{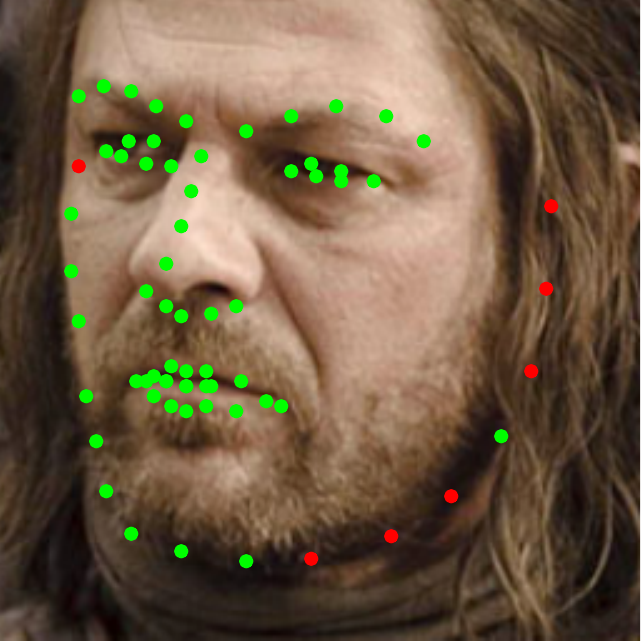}
\includegraphics[width=0.1\textwidth]{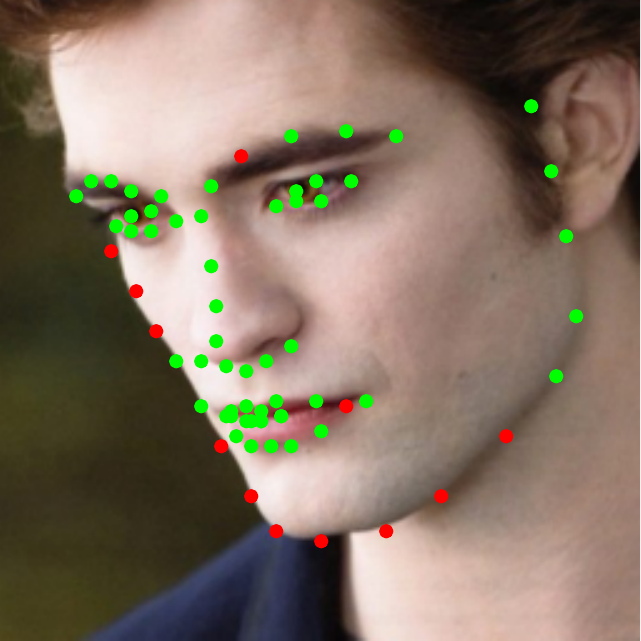}

\includegraphics[width=0.1\textwidth]{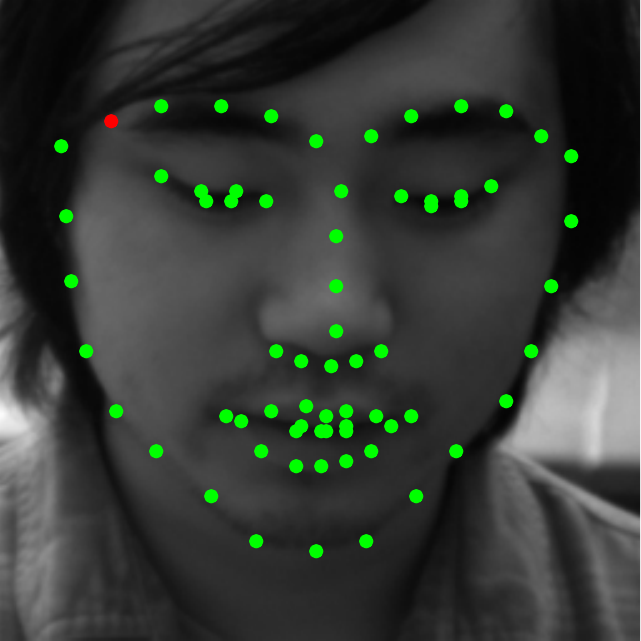}
\includegraphics[width=0.1\textwidth]{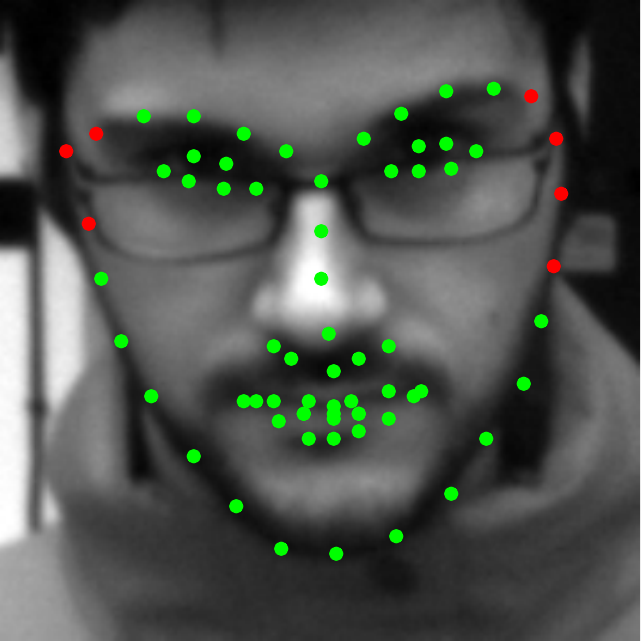}
\includegraphics[width=0.1\textwidth]{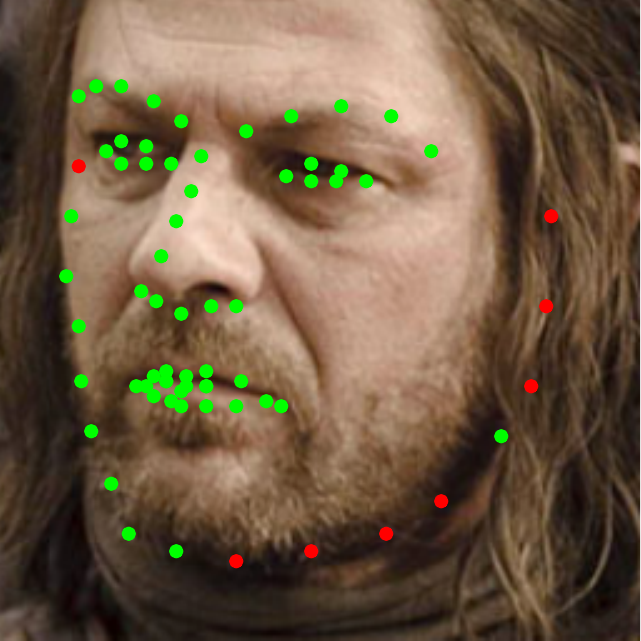}
\includegraphics[width=0.1\textwidth]{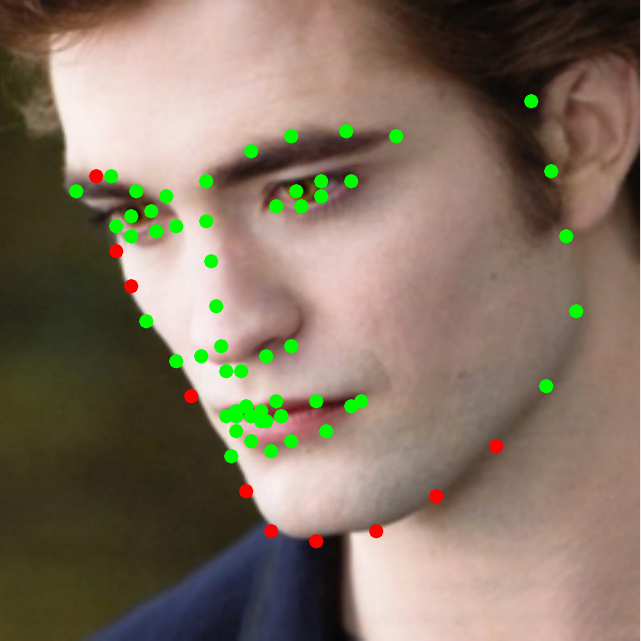}

\includegraphics[width=0.1\textwidth]{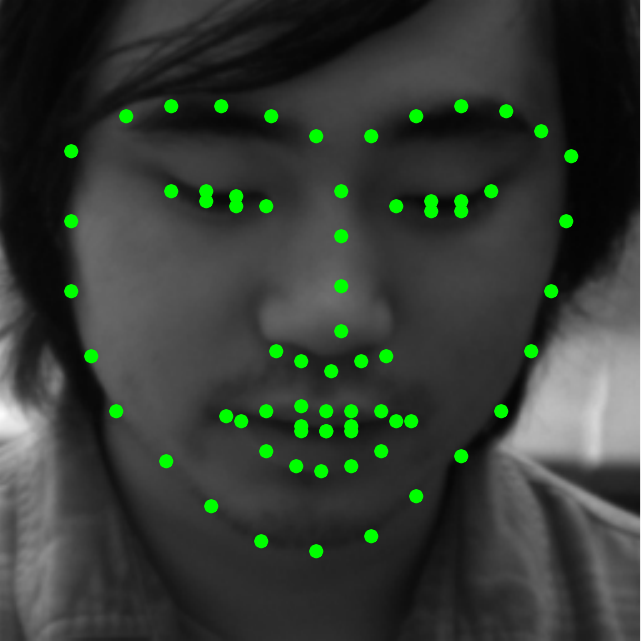}
\includegraphics[width=0.1\textwidth]{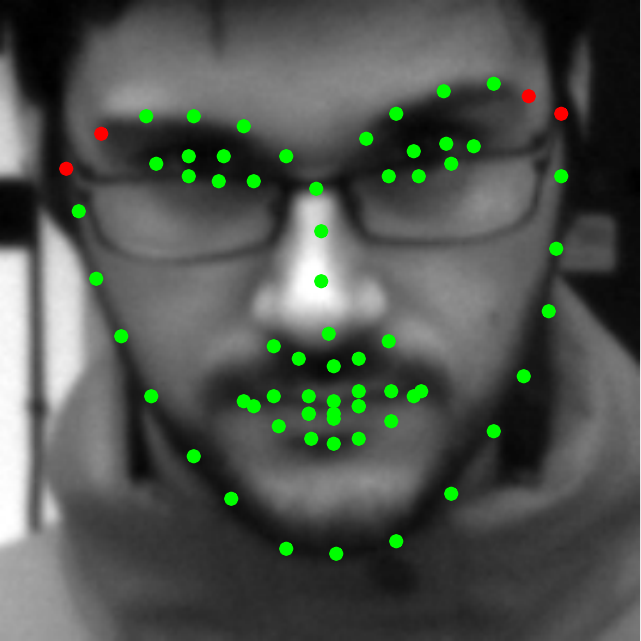}
\includegraphics[width=0.1\textwidth]{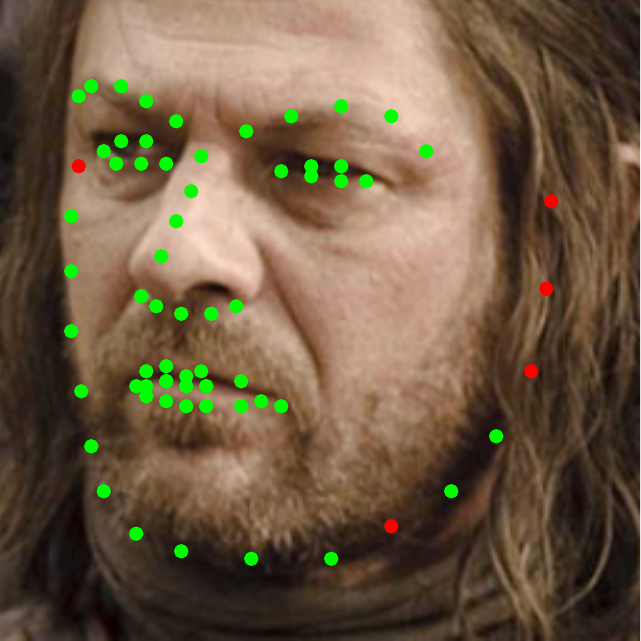}
\includegraphics[width=0.1\textwidth]{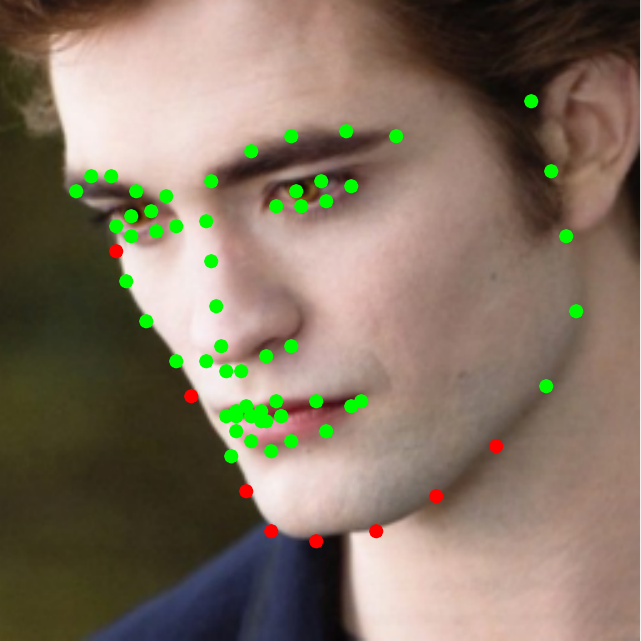}

\caption{Comparison of MobileFAN with different distillation method. Top row: Results of MobileFAN without distillation. Second row: Results of MobileFAN with feature-aligned distillation. Third row: Results of MobileFAN with feature-similarity distillation. Bottom row: Results of MobileFAN with both feature-aligned distillation and feature-similarity distillation. Red points indicate the normalized error is larger than $0.1$.}
\label{300w_ablation}
\vspace{-0.5cm}
\end{figure}

We can see from TABLE \ref{ablation_FA} that feature-aligned distillation improves the performance of our proposed ``MobileFAN". By utilizing the distillation loss generated from ``FA$_{3}$", ``MobileFAN" achieve $5.51\%$ mean error on $300$W Challenging subset.
Moreover, we notice that the performance can be further improved by adding more layers of distillation.
It is observed from TABLE \ref{ablation_FA} that ``MobileFAN + FA$_{3}$ + FA$_{2}$ +FA$_{1}$" achieves $2.00\%$ and $0.92\%$ relative improvement over ``MobileFAN + FA$_3$" and ``MobileFAN + FA$_2$ + FA$_3$", respectively.
Similarly, the mean error of ``MobileFAN + FA$_2$ + FA$_3$" has reduced from $5.51\%$ to $5.45\%$ compared with ``MobileFAN + FA$_3$".
Fig. \ref{300w_ablation} provides a straightforward comparison of MobileFAN without distillation and MobileFAN with feature-aligned distillation.

\begin{table}[t]
\setlength{\tabcolsep}{8pt}
\scriptsize
\centering
\caption{Mean error ($\%$) of feature-aligned distillation of different deconvolutional layers on $300$W Challenging subset.}
\vspace{-0.3cm}
\begin{tabular}{  l|c c }
\hline
 Method & MobileFAN \\
\hline
w/o distillation & $5.62$  \\
+ FA$_{3}$ & $5.51$\\
+ FA$_{3}$ + FA$_{2}$ & $5.45$\\
 + FA$_{3}$ + FA$_{2}$ +  FA$_{1}$ & $\mathbf{5.40}$\\
\hline
\end{tabular}
\label{ablation_FA}
\vspace{-0.3cm}
\end{table}

\begin{table}[t]
\setlength{\tabcolsep}{8pt}
\scriptsize
\centering
\caption{Mean error ($\%$) of feature-similarity distillation of different deconvolutional layers on $300$W Challenging subset.}
\vspace{-0.3cm}
\begin{tabular}{  l|c c }
\hline
 Method & MobileFAN \\
\hline
w/o distillation & $5.62$  \\
+ FS$_{3}$ & $5.52$\\
+ FS$_{3}$ + FS$_{2}$ & $5.47$\\
 + FS$_{3}$ + FS$_{2}$ +  FS$_{1}$ & $\mathbf{5.41}$\\
\hline
\end{tabular}
\label{ablation_FS}
\vspace{-0.7cm}
\end{table}

 \begin{figure}[ht]
\centering
\vspace{-0.2cm}
\includegraphics[width=8cm]{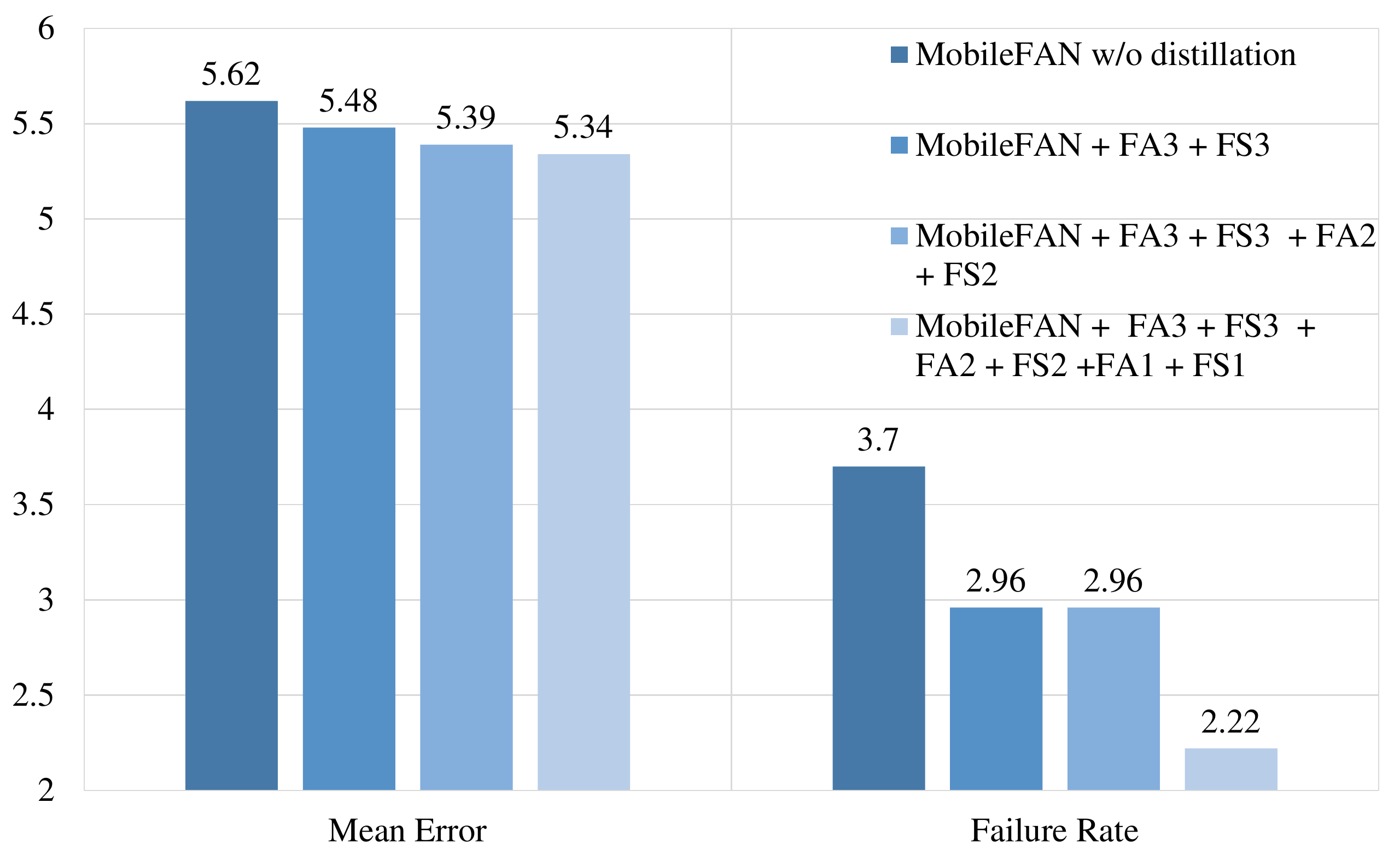}
\caption{Normalized mean error ($\%$) and failure rate ($\%$) on $300$W Challenging subset of various different combination of feature-aligned distillation and feature-similarity distillation.}
\label{ablation_bar_fr}
\vspace{-0.7cm}
\end{figure}

\noindent\textbf{Feature-similarity distillation over layer.}
To explore the effectiveness of feature-similarity distillation, we evaluate the mean error with various combinations of each layer on $300$W Challenging subset. As can be observed from TABLE \ref{ablation_FS}, applying more layers can lead to better performance.

In particular, ``MobileFAN + FS$_3$" outperforms ``MobileFAN" without distillation by a large margin, with a relative improvement of $1.78\%$ in mean error reduction. When one more layer of distillation is added, although the improvement is marginal, the mean error is reduced from $5.52\%$ to $5.47\%$. It is not surprising that ``MobileFAN (0.5) + FS$_{3}$ + FS$_{2}$ + FS$_{1}$" achieves the best performance of $5.41\%$ mean error among all versions of ``MobileFAN" on $300$W Challenging subset.
We can see a straightforward comparison in Fig. \ref{300w_ablation} that MobileFAN with feature-similarity distillation performs better than MobileFAN without distillation.

\noindent\textbf{Combination of feature-aligned distillation and feature-similarity distillation over layer.}
To evaluate the effectiveness of both using feature-aligned distillation and feature-similarity distillation, we report the normalized mean error and failure rate of various different combination of feature-aligned distillation and feature-similarity distillation in Fig. \ref{ablation_bar_fr}. It is shown that our ``MobileFAN" with knowledge distillation performs better when more layers of both feature-aligned distillation and feature-similarity distillation are used. In particular, ``MobileFAN + FA$_{3}$ + FS$_{3}$ + FA$_{2}$ + FS$_{2}$ + FA$_{1}$ + FS$_{1}$" performs better than ``MobileFAN + FA$_{3}$ + FS$_{3}$" and ``MobileFAN + FA$_{3}$ + FS$_{3}$ + FA$_{2}$ + FS$_{2}$", with a relative improvement of $2.55\%$ and $0.93\%$ in mean error reduction, respectively. we can also find that the failure rate of ``MobileFAN + FA$_{3}$ + FS$_{3}$" is lower than that of ``MobileFAN" without distillation. Similarly, the failure rate of ``MobileFAN + FA$_{3}$ + FS$_{3}$ + FA$_{2}$ + FS$_{2}$ + FA$_{1}$ + FS$_{1}$" drops from $2.96\%$ to $2.22\%$ compared with that of ``MobileFAN + FA$_{3}$ + FS$_{3}$ + FA$_{2}$ + FS$_{2}$".

 To summarize, the combination of feature-aligned distillation and feature-similarity distillation improves the performance of the compact network in facial landmark detection.

\noindent\textbf{The impact of $\lambda$.}
To investigate the impact of $\lambda$ on the training process of the proposed MobileFAN, we performed an ablation study of $\lambda$ regarding mean error on COFW dataset.
Experimental results in Table \ref{lambda} listed the mean errors obtained with different $\lambda$ ($\lambda$ increases from $10^{-5}$ to $10^{-1}$
).
$\lambda = 0$ means that the experiment is conducted on the proposed MobileFAN without distillation. 
It can be observed that 
the proposed MobileFAN achieved consistently lower mean errors with $\lambda$ varying from $10^{-5}$ to $10^{-2}$ compared with the mean error obtained with $\lambda=0$.
The best performance is achieved at $\lambda = 10^{-2}$ on COFW dataset.
Nevertheless, if $\lambda$ is too large (e.g. $\lambda = 10^{-1}$), the contribution from the supervision of the ground-truth heatmaps would be limited, and thus the proposed MobileFAN may fail to converge. The overall influence of $\lambda$ is positive to the training process of the proposed MobileFAN, indicating the effectiveness of the proposed distillation schemes.

\begin{table}[ht]
\small
\vspace{-0.2cm}
\centering
\caption{The influence of $\lambda$ on COFW dataset.}
\vspace{-0.1cm}
\begin{tabular}{ c |c |c |c |c |c |c }
\hline
$\lambda$ & $0$ & $10^{-5}$ & $10^{-4}$ & $10^{-3}$ & $10^{-2}$ & $10^{-1}$\\
\hline
Mean Error & 3.82 & 3.78 & 3.79 & 3.74 & 3.66 & 4.06\\

\hline
\end{tabular}
\label{lambda}
\vspace{-0.6cm}
\end{table}

\section{Conclusion}
In this paper, we focus on building a small facial landmark detection model, which remains an unsolved research problem. 
We propose a simple and lightweight \textit{Mobile Face Alignment Network} (MobileFAN) by using MobileNetV$2$ as the encoder and three simple deconvolutional layers as the decoder. 
This simple design significantly helps to reduce the computational burden.
With $11.5$ times fewer parameters compared with the state-of-the-art models, our MobileFAN still achieves comparable or even better performance on three challenging facial landmark detection datasets. A knowledge transfer technique is proposed to enhance the performance of MobileFAN. By transferring the finer structural information encoded by the teacher Network, the performance of the proposed MobileFAN is further improved in effectiveness for facial landmark detection.

{\small
\bibliographystyle{ieee_fullname}
\bibliography{face_alignment_V2}
}

\clearpage
\newpage

\end{document}